\documentclass[11pt, letterpaper]{article}

\usepackage{times}
\usepackage[margin=1.25in]{geometry}

\usepackage{amssymb, amsthm, bm, mathtools, amscd}
\usepackage[scr=boondoxo]{mathalfa}
\usepackage[usenames,dvipsnames,svgnames,table]{xcolor}
\usepackage[pdftex, xetex]{graphicx}
\usepackage[font={small}]{caption}
\usepackage{enumerate, setspace}
\usepackage{float, colortbl, tabularx, longtable, multirow, subcaption, environ, wrapfig, textcomp, booktabs}
\usepackage{pgf, tikz, framed}
\usepackage[normalem]{ulem}
\usetikzlibrary{arrows,positioning,automata,shadows,fit,shapes}
\usepackage[english]{babel}
\usepackage{blindtext}
\usepackage[toc,page,titletoc]{appendix}

\usepackage{microtype}
\microtypecontext{spacing=nonfrench}

\makeatletter
\def\th@plain{%
  \thm@notefont{}
  \itshape 
}
\def\th@definition{%
  \thm@notefont{}
  \normalfont 
}
\makeatother

\usepackage[colorlinks=true]{hyperref}
\usepackage{url, cite}

\newcommand\scalemath[2]{\scalebox{#1}{\mbox{\ensuremath{\displaystyle #2}}}}

\usepackage[capitalize]{cleveref}
\crefformat{equation}{(#2#1#3)}
\crefmultiformat{equation}{(#2#1#3)}%
{ and~(#2#1#3)}{, (#2#1#3)}{ and~(#2#1#3)}
\crefformat{theorem}{Theorem~#2#1#3}
\crefformat{lemma}{Lemma~#2#1#3}
\crefformat{remark}{Remark~#2#1#3}
\crefformat{section}{Section~#2#1#3}
\crefformat{assumption}{Assumption~#2#1#3}
\crefformat{example}{Example~#2#1#3}
\crefrangeformat{section}{Sections~#3#1#4--#5#2#6}
\crefformat{assumption}{Assumption~#2#1#3}
\crefformat{example}{Example~#2#1#3}

\crefrangeformat{equation}{~(#3#1#4--#5#2#6)}
\crefrangeformat{figure}{Figures~#3#1#4--#5#2#6}
\crefrangeformat{assumption}{Assumptions~(#3#1#4--#5#2#6)}

\renewcommand{\qed}{\hfill \mbox{\raggedright \rule{0.1in}{0.1in}}}

\newcommand{\tbf}[1]{\textbf{#1}}

\DeclarePairedDelimiter{\norm}{\lVert}{\rVert}

\DeclareMathOperator*{\argmin}{\textrm{argmin}}

\newcommand{\aed}[1]{\begin{aligned} #1 \end{aligned}}
\newcommand{\beq}[1]{\begin{equation}#1\end{equation}}
\newcommand{\beqs}[1]{\begin{equation*}#1\end{equation*}}

\newcommand{\trm}[1]{\mathrm{#1}}

\newcommand{\pmat}[1]{\begin{pmatrix}#1\end{pmatrix}}

\providecommand\f[2]{\ensuremath \frac{#1}{#2}}

\providecommand\f[2]{\ensuremath \frac{#1}{#2}}
\providecommand\rbrac[1]{\ensuremath \left(#1\right)}

\providecommand\sqbrac[1]{\ensuremath \left[#1\right]}
\providecommand\Sqbrac[1]{\ensuremath \big[#1 \big]}

\providecommand\cbrac[1]{\ensuremath \left\{#1\right\}}

\providecommand\ag[1]{\ensuremath \left\langle#1\right\rangle}

\newcommand{\grad}{\nabla}

\theoremstyle{plain}
\newtheorem{theorem}{Theorem}

\newtheorem{lemma}[theorem]{Lemma}

\theoremstyle{definition}

\newtheorem{remark}[theorem]{Remark}

\makeatletter

\makeatletter

\DeclareMathOperator*{\E}{\mathbb{E}}

\providecommand{\ind}{{\bf 1}}

\definecolor{color_skyblue}{rgb}{0.01,0.39,0.75}

\def \s {\sigma}

\def \th {\theta}
\def \a {\alpha}
\def \b {\beta}
\def \p {\varphi}

\def \g {\gamma}

\def \l {\lambda}
\def \k {\kappa}

\def \G {\Gamma}

\def \Th {\Theta}

\def \BB {\mathcal{B}}

\def \BB {\mathcal{B}}

\def \reals {\mathbb{R}}

\newcommand{\ignore}[1]{}

\newcommand{\heading}[1]{\smallskip \noindent \tbf{#1.}}
\def \d {\trm{d}}

\def \dx {\d x}
\def \dy {\d y}
\def \dz {\d z}

\newcommand {\df}[2][] {\f{\trm{d} #1}{\trm{d} #2}}

\newcommand {\pf}[2][] {\f{\partial #1}{\partial #2}}
\newcommand {\ppf}[2][] {\f{\partial^2 #1}{\partial #2^2}}
\newcommand {\ppftwo}[2][] {\f{\partial^2 #1}{#2}}

\def \ff {F}

\def \mm {m(z)}
\def \mmth {m_\th(z)}

\def \ee {e(z | x)}
\def \eeth {e_\th(z | x)}
\def \eethk {e_{\th^k}(z | x)}

\def \dd {d(x | z)}
\def \ddth {d_\th(x | z)}

\def \cc {c(y | z)}
\def \ccth {c_\th(y | z)}
\def \ccxth {c_\th(y_x | z)}

\def \kl {\trm{KL}}

\def \zthx {Z_{\th, x}}

\def \thlg {\Th_{\l, \g}}

\def \jthlg {J(\th, \l, \g)}

\def \ldot {\dot{\l}}
\def \gdot {\dot{\g}}
\def \thdot {\dot{\th}}

\def \ww {W_2}

\def \hthl {\hat{\th}_\l}
\def \htht {\hat{\th}_t}

\hypersetup{
    citecolor=blue
}

\usepackage[sectionbib,square]{natbib}

\def \titlee {A Free-Energy Principle for Representation Learning}

\title{
\tbf{\titlee}}
\date{}

\author{Yansong Gao$^{1}$ and Pratik Chaudhari$^{2}$\\[0.1in]
\small $^{1}$Applied Mathematics and Computational Science, University of Pennsylvania.\\
\small $^{2}$Department of Electrical and Systems Engineering, University of Pennsylvania.\\
\small Email: \href{mailto:gaoyans@sas.upenn.edu}{gaoyans@sas.upenn.edu},
       \href{mailto:pratikac@seas.upenn.edu}{pratikac@seas.upenn.edu}%
}
\graphicspath{{./fig/}}

\begin{document}
\maketitle

\begin{abstract}
This paper employs a formal connection of machine learning with thermodynamics to characterize the quality of learnt representations for transfer learning. We discuss how information-theoretic functionals such as rate, distortion and classification loss of a model lie on a convex, so-called equilibrium surface. We prescribe dynamical processes to traverse this surface under constraints, e.g., an iso-classification process that trades off rate and distortion to keep the classification loss unchanged. We demonstrate how this process can be used for transferring representations from a source dataset to a target dataset while keeping the classification loss constant. Experimental validation of the theoretical results is provided on standard image-classification datasets.\\

\noindent \tbf{Keywords:} information theory; thermodynamics; rate-distortion theory; transfer learning; information bottleneck; optimal transportation
\end{abstract}

\section{Introduction}
\label{s:intro}

A representation is a statistic of the data that is ``useful''. Classical Information Theory creates a compressed representation and makes it easier to store or transmit data; the goal is always to decode the representation to get the original data back.
If we are given images and their labels, we could learn a representation that is useful to predict the correct labels. This representation is thus a statistic of the data \emph{sufficient} for the task of classification. If it is also minimal---say in its size---it would discard information in the data that is not correlated with the labels.
Such a representation is unique to the chosen task, it would perform poorly to predict some other labels correlated with the discarded information. If instead the representation were to have lots of redundant information about the data, it could potentially predict other labels correlated with this extra information.

The premise of this paper is our desire to characterize the information discarded in the representation when it is fit on a task. We want to do so in order to learn representations that can be transferred easily to other tasks.

Our main idea is to choose a canonical task---in this paper, we pick reconstruction of the original data---as a way to measure the discarded information. Although one can use any canonical task, reconstruction is special. It is a ``capture all'' task in the sense that achieving perfect reconstruction entails that the representation is lossless; information discarded by the original task is therefore readily measured as the one that helps solve the canonical task. This leads to the study of the following Lagrangian which is similar to the Information Bottlenck of~\citet{tishby2000information}
\beqs{
    \ff(\l,\g) = \min_{\substack{\th \in \Th, \eeth, \mmth,\\ \ddth, \ccth}}\ R + \l D + \g C
    \label{eq:F_intro}
}
where the rate $R$ is an upper bound on the mutual information of the representation learnt by the encoder $\eeth$ with the input data $x$, distortion $D$ measures the quality of reconstruction of the decoder $\ddth$ and $C$ measures the classification loss of the classifier $\ccth$. As~\citet{alemi2018therml} show, this Lagrangian can be formally connected to ideas in thermodynamics. We heavily exploit and specialize this point of view, as summarized next.

\subsection{Summary of contributions}

Our main technical observation is that $\ff(\l,\g)$ can be intepreted as a free-energy and a stochastic learning process that minimizes its corresponding Hamiltonian converges to the optimal free-energy. This corresponds to an ``equilibrium surface'' of information-theoretic functionals $R, D$ and $C$ and a surface $\thlg$ of the model parameters at convergence. We prove that the equilibrium surface is convex and its dual, the free-energy $\ff(\l, \g)$, is concave. The free-energy is only a function of Lagrange multipliers $(\l, \g)$, the family of model parameters $\Th$, and the task, and is therefore invariant of the learning dynamics.

Second, we design a quasi-static stochastic process, akin to an equilibrium process in thermodynamics, to keep the model parameters $\th$ on the equilibrium surface. Such a process allow us to travel to any feasible values of $(R,D,C)$ while ensuring that the parameters $\th$ of the model are on the equilibrium surface. We focus on one process, the ``iso-classification process'' which automatically trades off the rate and distortion to keep the classification loss constant.

We prescribe a quasi-static process that allows for a controlled transfer of learnt representations. It adapts the model parameters as the task is changed from some source dataset to a target dataset while keeping the classification loss constant. Such a process is in stark contrast to current techniques in transfer learning which do not provide any guarantees on the quality of the model on the target dataset.

We provide extensive experimental results which realize the theory developed in this paper.

\section{Theroetical setup}
\label{s:prelim}

This section introduces notation and preliminaries that form the building blocks of our approach.

\subsection{Auto-Encoders}
\label{ss:prelim:auto_encoder}

Consider an encoder $\ee$ that encodes data $x$ into a latent code $z$ and a decoder $\dd$ that decodes $z$ back into the original data $x$. If the true distribution of the data is $p(x)$ we may define the following functionals.
\beq{
    \aed{
        H &= \E_{x \sim p(x)} \Sqbrac{-\log p(x)}\\
        D &= \E_{x \sim p(x)} \sqbrac{-\int \dz\ \ee \log \dd}\\
        R &= \E_{x \sim p(x)} \sqbrac{\int \dz\ \ee \log \f{\ee}{\mm}}
    }
    \label{eq:hdr}
}
We denote expectation over data using the notation $\ag{\p}_{p(x)} = \int \dx\ p(x) \p$. The first functional $H$ is the Shanon entropy of the true data distribution; it quantifies the complexity of the data. The distortion $D$ measures the quality of the reconstruction through its log-likelihood. The rate $R$ is a Kullback-Leibler (KL) divergence; it measures the average excess bits used to encode samples from $\ee$ using a code that was built for our approximation of the true marginal on the latent factors $\mm$.

\subsection{Rate-Distortion curve}
\label{ss:prelim:rate_distortion}

\begin{figure}[!htpb]
\centering
    \begin{subfigure}[b]{0.3\textwidth}
    \centering
    \includegraphics[width=\textwidth]{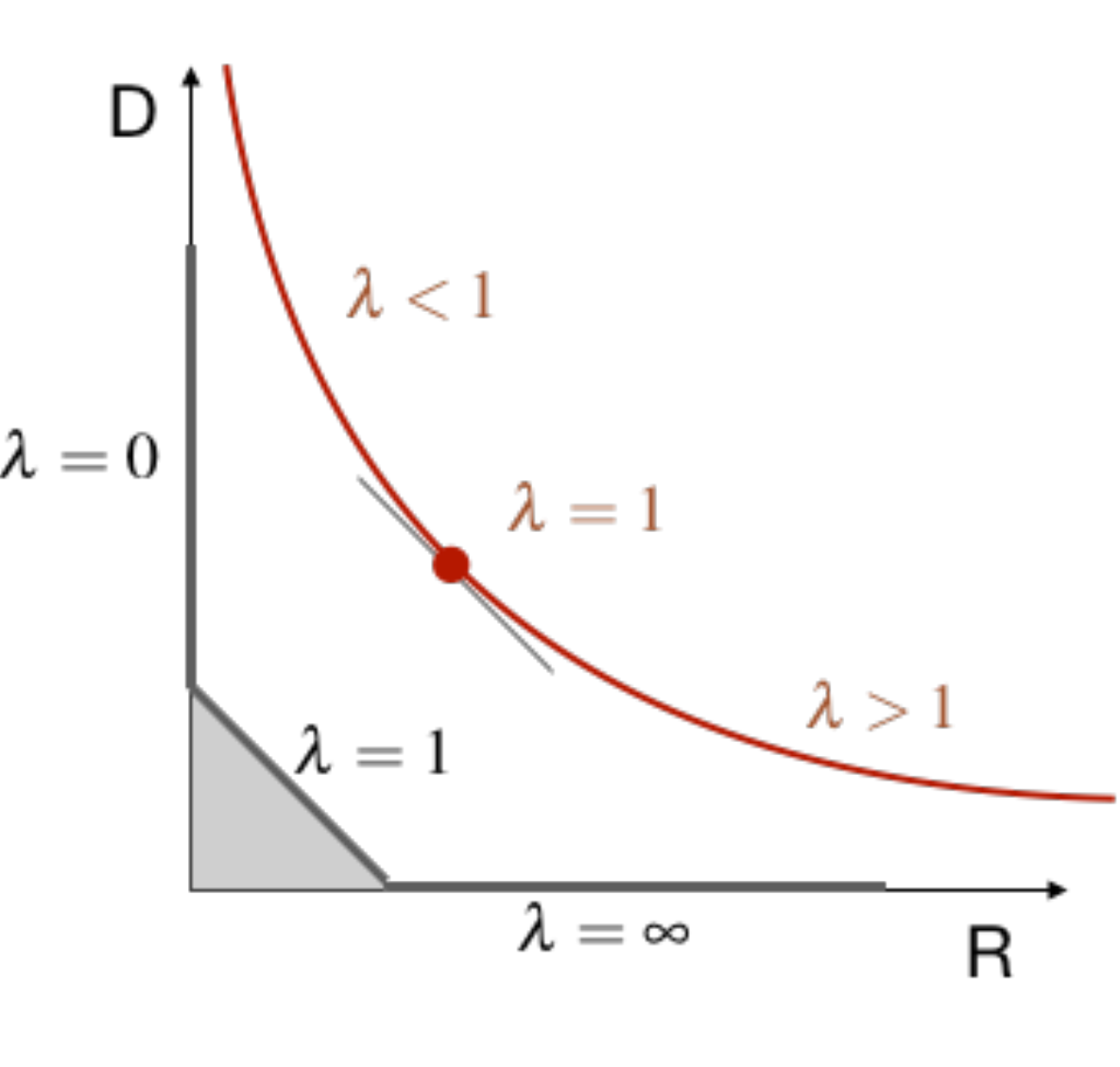}
    \caption{}
    \label{fig:rd}
    \end{subfigure}
    \hspace{0.3in}
    \begin{subfigure}[b]{0.3\textwidth}
    \centering
    \includegraphics[width=\textwidth]{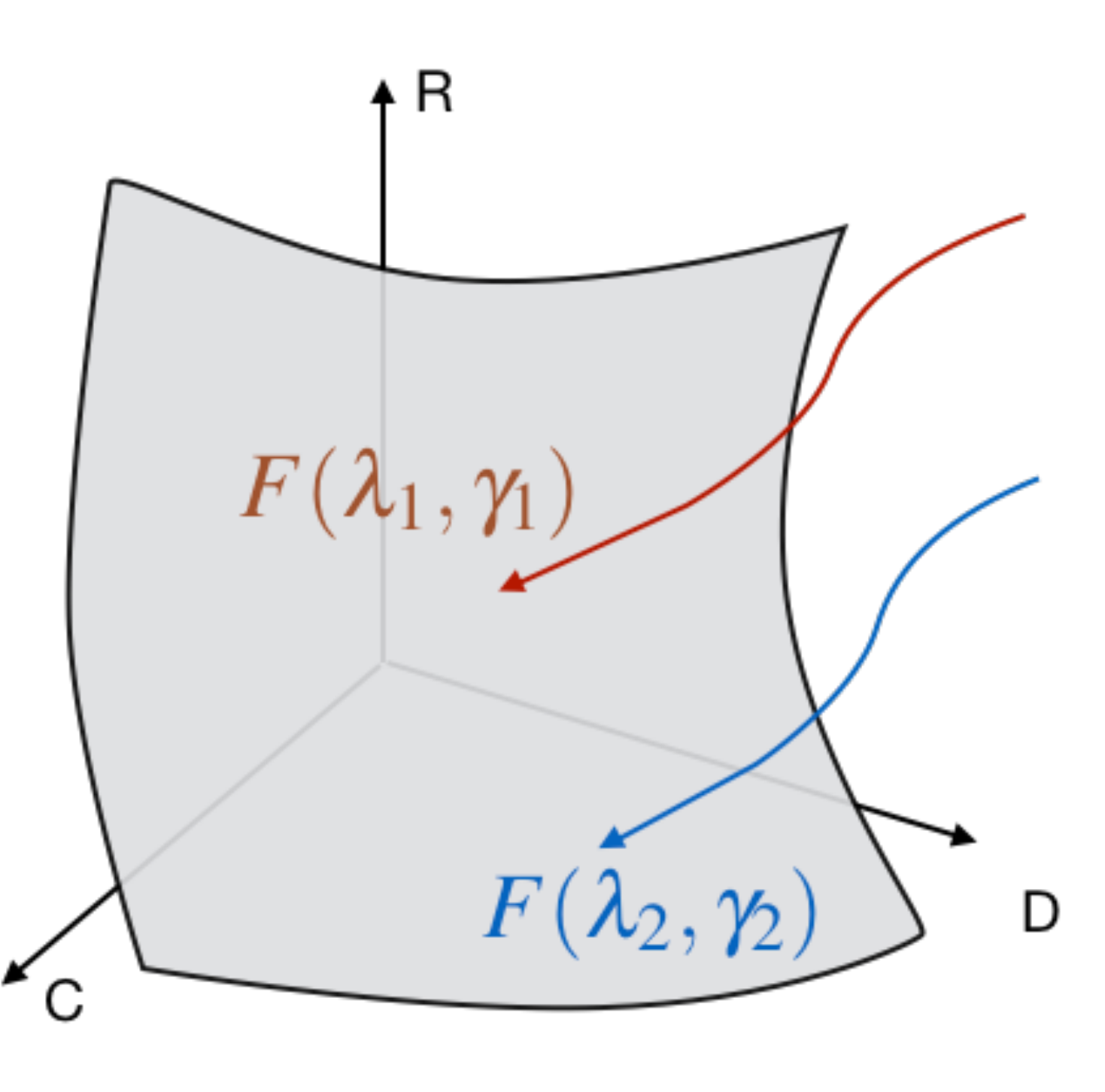}
    \caption{}
    \label{fig:rdc}
    \end{subfigure}
\caption{\tbf{Schematic of the equilibrium surface.} \cref{fig:rd} shows that rate ($R$) and distortion ($D$) trade off against each other on the equilibrium surface. Similarly in~\cref{fig:rdc}, the equilibrium surface is a convex constraint that joins rate, distortion and the classification loss. Training objectives with different $(\l, \g)$ (shown in red and blue) reach different parts of the equilibrium surface.}
\label{fig:rd_rdc}
\end{figure}

The functionals in~\cref{eq:hdr} come together to give the inequality
\beq{
    H - D \leq I_e(x; z) \leq R
    \label{eq:hdr_ineq}
}
where $I_e = \kl(\ee\ ||\ p(z|x))$ is the KL-divergence between the learnt encoder and the true (unknown) conditional of the latent factors.
The outer inequality $H \leq D + R$ forms the basis for a large body of literature on Evidence Lower Bounds (ELBO, see~\citet{kingma2013auto}). Consider~\cref{fig:rd}, if the capacity of our candidate distributions $\ee, \mm$ and $\dd$ is infinite, we can obtain the equality $H = R + D$. This is the thick black line in~\cref{fig:rd}.

For finite capacity variational families, say parameterized by $\th$, which we denote by $\eeth$, $\ddth$ and $\mmth$ respectively, as~\citet{alemi2017fixing} argue, one obtains a convex RD curve (shown in red in~\cref{fig:rd}) corresponding to the Lagrangian
\beq{
    \ff(\l) = \min_{\eeth, \mmth, \ddth}\ R + \l D.
    \label{eq:f_rd}
}

This Lagrangian is the relaxation of the idea that given a fixed variational family and data distribution $p(x)$, there exists an optimal value of, say, rate $R = f(D)$ that best sandwiches~\cref{eq:hdr_ineq}. The optimal Lagrange multiplier is $\l = \pf[R]{D}$ evaluated at the desired value of $D$.



\subsection{Incorporating the classification loss}
\label{ss:prelim:classification}

Let us create a classifier that uses the learnt representation $z$ as the input and set the classification loss as the negative log-likelihood of the prediction
\beq{
    C = \E_{x \sim p(x)}\sqbrac{-\int\ \dz\ \ee \log \cc}.
    \label{eq:c}
}
If the parameters of the model---which now consists of the encoder $\ee$, decoder $\dd$ and the classifier $\cc$---are denoted by $\th$, the training process for the model induces a distribution $p(\th|\cbrac{(x,y)})$ where $\cbrac{(x,y)}$ denotes a finite dataset. In addition to $R, D$ and $C$, the authors in~\citet{alemi2018therml} define
\beq{
    S = \E_{x \sim p(x), y \sim p(y|x)} \sqbrac{\log \f{p(\th|\cbrac{x,y})}{m(\th)}}
    \label{eq:s}
}
which is the relative entropy of the distribution on parameters $\th$ after training compared to a prior distribution $m(\th)$ of our choosing.
Using a very similar argument as~\cref{ss:prelim:rate_distortion} the four functionals $R, D, C$ and $S$ form a convex three-dimensional surface in the RDCS phase space. A schematic is shown in~\cref{fig:rdc} for $\s = 0$. We can again consider a Lagrange relaxation of this surface given by
\beq{
    \ff(\l, \g, \s) = \min_{\ee, \mm, \dd, \cc}\ R + \l D + \g C + \s S.
    \label{eq:f_rdcs}
}

\begin{remark}[`The `First Law'' of learning]
\label{rem:first_law}
\citet{alemi2018therml} draw formal connections of the Lagrangian in~\cref{eq:f_rdcs} with the theory of thermodynamics. Just like the first law of thermodynamics is a statement about the conservation of energy in physical processes, the fact that the four functionals are tied together in a smooth constraint $f(R, D, C, S) = 0$ leads to an equation of the form
\beq{
    \d R = - \l\ \d D - \g\ \d C - \s\ \d S
    \label{eq:first_law}
}
which indicates that information in learning processes is conserved. The information in the latent representation $z$ is kept either to reconstruct back the original data or to predict the labels. The former is captured by the encoder-decoder pair, the latter is captured by the classifier.
\end{remark}

\begin{remark}[Setting $\s = 0$]
\label{rem:s_zero}
The distribution $p(\th | \cbrac{(x,y)})$ is a posterior on the parameters of the model given the dataset. While this distribution is well-defined under minor technical conditions, e.g., ergodicity, performing computations with this distribution is difficult.
\tbf{We therefore only consider the case when $\s = 0$ in the sequel} and leave the general case for future work.
\end{remark}

The following lemma (proved in~\cref{s:app:proof_r_cvx_f_noncvx}) shows that the constraint surface connecting the information-theoretic functionals $R, D$ and $C$ is convex and its dual, the Lagrangian $F(\l, \g)$ is concave.
\begin{lemma}[The $RDC$ constraint surface is convex]
\label{lem:r_cvx_f_noncvx}
The constraint surface $f(R, D, C) = 0$ is convex and the Lagrangian $\ff(\l,\g)$ is concave.
\end{lemma}
We can show using a similar proof that the entire surface joining $R, D, C$ and $S$ is convex by considering the cases $\l = 0$ and $\g =0$ separately. Note that the constraint is convex in $R, D$ and $C$; it need not be convex in the model parameters $\th$ that parameterize $\eeth, \mmth$, etc.

\subsection{Equilibrium surface of optimal free-energy}
\label{ss:prelim:f}

We next elaborate upon the objective in~\cref{eq:f_rdcs}. Consider the functionals $R, D$ and $C$ parameterized using parameters $\th \in \Th \subseteq \reals^N$. First, consider the problem
\beq{
    \ff(\l, \g) = \min_{\ee,\ \th \in \Th}\ R + \l D + \g C.
    \label{eq:f_eth_th}
}
We can solve this using calculus of variations to get
\[
    \ee \propto \mmth\ \ddth^\l \exp \rbrac{\g \int\ \dy\ p(y|x)\ \log \ccth}.
\]
We assume in this paper that the labels are a deterministic function of the data, i.e., $p(y|x) = \delta(y - y_x)$ where $y_x$ is the true label of the datum $x$. We therefore have
\[
    \ee = \f{\mmth \ddth^\l \ccxth^\g}{\zthx}
\]
where the normalization constant is
\beq{
    \zthx = \int\ \dz\ \mmth \ddth^\l \ccxth^\g.
    \label{eq:z}
}
The objective $\ff(\l,\g)$ can now be rewritten as maximizing the log-partition function, also known as the free-energy in statistical physics~\citep{mezard2009information},
\beq{
    \ff(\l, \g) = \min_{\th \in \Th}\ -\ag{\log \zthx}_{p(x)}.
    \label{eq:f}
}

\begin{remark}[Why is it called the ``equilibrium'' surface?]
\label{rem:why_equilibrium}
Given a finite dataset $\cbrac{(x,y)}$, one may minimize the objective in~\cref{eq:f_eth_th} using stochastic gradient descent (SGD,~\citet{robbins1951stochastic}) on a Hamiltonian
\beq{
    H(z; x, \th, \l, \g) \equiv -\log \mmth - \l \log \ddth - \g \log \ccth
    \label{eq:H}
}
with updates given by
\beq{
    \th^{k+1} = \th^k - \s\ \grad_\th \E_{x \sim p(x)} \sqbrac{\int\ \dz\ \eethk H(z; x, \th^k, \l, \g)}
    \label{eq:sgd_H}
}
where $\s > 0$ is the step-size; the gradient $\grad_\th$ is evaluated over samples from $p(x)$ and $\eeth$.
Using the same technique as that of~\citet{chaudhari2017stochastic}, one can show that the objective
\[
    \E_{\th \sim p(\th | \cbrac{x,y})}\ \Sqbrac{\ag{-\log \zthx}_{p(x)}} - \s H(p(\th\ |\ \cbrac{x,y})).
\]
decreases \emph{monotonically}. Observe that our objective in~\cref{eq:f_eth_th} corresponds to the limit $\s \to 0$ of this objective along with a uniform non-informative prior $m(\th)$ in~\cref{eq:s}. In fact, this result is analogous to the classical result that an ergodic Markov chain makes monotonic improvements in the KL-divergence as it converges to the steady-state, also known as, equilibrium, distribution~\citep{levin2017markov}. The posterior distribution of the model parameters induced by the stochastic updates in~\cref{eq:sgd_H} is the Gibbs distribution $p^*(\th\ |\ \cbrac{(x,y)}) \propto \exp \rbrac{-2(R + \l D + \g C)/\s}$.

It is for the above reason that we call the surface in~\cref{fig:rdc} parameterized by
\beq{
    \thlg = \cbrac{\th \in \Th: -\ag{\log \zthx}_{p(x)} = \ff(\l, \g)}
    \label{eq:thlg}
}
as the ``equilibrium surface''. Learning, in this case minimizing~\cref{eq:f_eth_th}, is initialized outside this surface and converges to specific parts of the equilibrium surface depending upon $(\l, \g)$; this is denoted by the red and blue curves in~\cref{fig:rdc}. The constraint that ties results in this equilibrium surface is that variational inequalities such as~\cref{eq:hdr_ineq} (more are given in~\citet{alemi2018therml}) are tight up to the capacity of the model. This is analogous to the concept of equilibrium in thermodynamics~\citep{sethna2006statistical}
\end{remark}

\section{Dynamical processes on the equilibrium surface}
\label{s:dynamical_processes_equilibrium_surface}

This section prescribes dynamical processes that explore the equilibrium surface.
For any parameters $\th \in \Th$, not necessarily on the equilibrium surface, let us define
\beq{
    \jthlg = -\ag{\log \zthx}_{p(x)}.
    \label{eq:j}
}
If $\th \in \thlg$ we have $\jthlg = \ff(\l,\g)$ which implies
\beq{
    \grad_\th \jthlg = 0\ \textrm{for all}\ \th \in \thlg.
    \label{eq:quasi_static}
}
\paragraph{Quasi-static process.} A quasi-static process in thermodynamics happens slowly enough for a system to remain in equilibrium with its surroundings. In our case, we are interested in evolving Lagrange multipliers $(\l, \g)$ slowly and simultaneously keep the model parameters $\th$ on the equilibrium surface; the constraint~\cref{eq:quasi_static} thus holds at each time instant. The equilibrium surface is parameterized by $R, D$ and $C$ so changing $(\l, \g)$ adapts the three functionals to track their optimal values corresponding to $\ff(\l, \g)$.

Let us choose some values $(\ldot, \gdot)$ and the trivial dynamics $\df{t} \l = \ldot$ and $\df{t} \g = \gdot$. The quasi-static constraint leads to the following partial differential equation (PDE)
\beq{
    0 \equiv \df{t} \grad_\th \jthlg = \grad_\th^2 J\ \thdot + \ldot \pf{\l} \grad_\th J + \gdot \pf{\g} \grad_\th J
    \label{eq:quasi_static_pde}
}
valid all $\th \in \thlg$. At each location $\th \in \thlg$ the above PDE indicates how the parameters should evolve upon changing the Lagrange multipliers $(\l, \g)$. We can rewrite the PDE using the Hamiltonian $H$ in~\cref{eq:H} as shown next.

\begin{lemma}[Equilibrium dynamics for parameters $\th$]
\label{lem:equilibrium_dynamics}
Given $(\ldot, \gdot)$, the parameters $\th \in \thlg$ evolve as
\beq{
    \aed{
        \thdot &= A^{-1} b_\l\ \ldot + A^{-1} b_\g\ \gdot \\
        &= \th_\l \ldot + \th_\g \gdot
    }
    \label{eq:th_equilibrium_dynamics}
}
where $H$ is the Hamiltonian in~\cref{eq:H} and
\[
    \aed{
    A   &= \grad_\th^2 J = \E_{x \sim p(x)} \sqbrac{ \ag{\grad_\th^2 H} + \ag{\grad_\th H} \ag{\grad_\th H}^\top - \ag{\grad_\th H\ \grad_\th^\top H}};\\
    b_\l &=  -\pf{\l} \grad_\th J = \scalemath{1}{-\E_{x \sim p(x)} \sqbrac{ \ag{\pf[\grad_\th H]{\l}} - \ag{\pf[H]{\l} \grad_\th H} + \ag{\pf[H]{\l}} \ag{\grad_\th H} }};\\
    b_\g &=  -\pf{\g} \grad_\th J = \scalemath{1}{-\E_{x \sim p(x)} \sqbrac{ \ag{\pf[\grad_\th H]{\g}} - \ag{\pf[H]{\g} \grad_\th H} + \ag{\pf[H]{\g}} \ag{\grad_\th H} }}.
    }
\]
All the inner expectations $\ag{\cdot}$ above are taken with respect to the Gibbs measure of the Hamiltonian, i.e., $\ag{\p} = \f{\int\ \p \exp(-H(z))\ \dz}{\int\ \exp(-H(z))\ \dz}$. The dynamics for the parameters $\th$ is therefore a function of the two directional derivatives
\beq{
    \th_\l =  A^{-1}\ b_\l,\quad\textrm{and}\quad \th_\g =  A^{-1}\ b_\g
    \label{eq:thl_thg}
}
with respect to $\l$ and $\g$. Note that $A$ in~\cref{eq:th_equilibrium_dynamics} is the Hessian of a strictly convex functional.
\end{lemma}

This lemma allows us to implement dynamical processes for the model parameters $\th$ on the equilibrium surface. As expected, this is an ordinary differential equation~\cref{eq:th_equilibrium_dynamics} that depends on our chosen evolution for $(\ldot, \gdot)$ through the directional derivatives $\th_\l, \th_\g$. The utility of the above lemma therefore lies in the expressions for these directional derivatives. \cref{s:app:proof_equilibrium_dynamics} gives the proof of the above lemma.

\begin{remark}[Implementing the equilibrium dynamics]
\label{rem:implementing_lem_eqdyn}
The equations in~\cref{lem:equilibrium_dynamics} may seem complicated to compute but observe that they can be readily estimated using samples from the dataset $x \sim p(x)$ and those from the encoder $z \sim \eeth$. The key difference between~\cref{eq:th_equilibrium_dynamics} and, say, the ELBO objective is that the gradient in the former depends upon the Hessian of the Hamiltonian $H$. These equations can be implemented using Hessian-vector products~\citep{pearlmutter1994fast}. If the dynamics involves certain constrains among the functionals, as~\cref{rem:implementing_iso_c} shows, we simplify the implementation of such equations.
\end{remark}

\subsection{Iso-classification process}
\label{ss:iso_classification}

An iso-thermal process in thermodynamics is a quasi-static process where a system exchanges energy with its surroundings and remains in thermal equilibrium with the surroundings. We now analogously define an iso-classification process that adapts parameters of the model $\th$ while the free-energy is subject to slow changes in $(\l, \g)$. This adaptation is such that the classification loss is kept constant while the rate and distortion change automatically.

Following the development in~\cref{lem:equilibrium_dynamics}, it is easy to create an iso-classification process. We simply add a constraint of the form
\beq{
    \aed{
        \df{t} \grad_\th J = 0 &\qquad \textrm{(Quasi-Static Condition)}\\
        \df{t} C = 0           &\qquad \textrm{(Iso-classification Condition)}.
    }
    \label{eq:iso_c_conditions}
}
Using a very similar computation (given in~\cref{s:app:iso_c_constraint}) as that in the proof of~\cref{lem:equilibrium_dynamics}, this leads to the constrained dynamics
\beq{
    \aed{
        0 &= C_\l \ldot + C_\g \gdot\\
        \thdot &= \th_\l \ldot + \th_\g \gdot.
    }
    \label{eq:iso_c_constraint}
}
The quantities $C_\l$ and $C_\g$ are given by
\beq{
    \aed{
        C_\l &=
            \scalemath{1}{
            -\E_{x \sim p(x)} \sqbrac{
            \ag{\pf[H]{\l}} \ag{\ell} - \ag{\pf[H]{\l}\ \ell} + \ag{\th_\l^{\top}  \grad_\th H} \ag{\ell} - \ag{ \ell \th_\l^{\top} \grad_\th H } + \ag{\th_\l^{\top} \grad_\th \ell} }}
            \\
        C_\g &=
            \scalemath{1}{
            -\E_{x \sim p(x)} \sqbrac{
            \ag{\pf[H]{\g}} \ag{\ell} - \ag{\pf[H]{\g}\ \ell} +
            \ag{\th_\g^{\top} \grad_\th H} \ag{\ell} - \ag{\ell \th_\g^{T} \grad_\th H } + \ag{\th_\g^{\top} \grad_\th \ell} }}
    }
    \label{eq:cl_cg}
}
where $\ell = \log \ccxth$ is the logarithm of the classification loss. Observe that we are not free to pick any values for $(\ldot, \gdot)$ for the iso-classification process anymore, the constraint $\df[C]{t} = 0$ ties the two rates together.

\begin{remark}[Implementing an iso-classification process]
\label{rem:implementing_iso_c}
The first constraint in~\cref{eq:iso_c_constraint} allows us to choose
\beq{
    \aed{
        \ldot &= -\a \pf[C]{\g} = -\a \ppf[F]{\g}\\
        \gdot &= \a \pf[C]{\l} = \a \f{\partial^2 F}{\partial \l \partial \g}
    }
    \label{eq:dl_dg_isoc}
}
where $\a$ is a parameter to scale time. The second equalities in both rows follow because $F(\l, \g)$ is the optimal free-energy which implies relations like $D = \pf[F]{\l}$ and $C = \pf[F]{\g}$. We can now compute the two deriatives in~\cref{eq:dl_dg_isoc} using finite differences to implement an iso-classification process. This is equivalent to running the dynamics in~\cref{eq:iso_c_constraint} using finite-difference approximation for the terms $\pf[H]{\l}$, $\pf[H]{\g}$, $\pf[\grad_\th H]{\l}$, $\pf[\grad_\th H]{\g}$. While approximating all these listed quantities at each update of $\th$ would be cumbersome, exploiting the relations in~\cref{eq:iso_c_constraint} is efficient even for large neural networks, as our experiments show.
\end{remark}

\begin{remark}[Other dynamical processes of interest]
\label{rem:other_processes}
In this paper, we focus on iso-classification processes. However, following the same program as that of this section, we can also define other processes of interest, e.g., one that keeps $C + \b^{-1} R$ constant while fine-tuning a model. This is similar to the alternative Information Bottleneck of~\citet{achille2017emergence} wherein the rate is defined using the weights of a network as the random variable instead of the latent factors $z$. This is also easily seen to be the right-hand side of the PAC-Bayes generalization bound~\citep{mcallester2013pac}. A dynamical process that preserves this functional would be able to control the generalization error which is an interesting prospect for future work.
\end{remark}

\section{Transferring representations to new tasks}
\label{s:transfer}

\cref{s:dynamical_processes_equilibrium_surface} demonstrated dynamical processes where the Lagrange multipliers $\l, \g$ change with time and the process adapts the model parameters $\th$ to remain on the equilibrium surface. This section demonstrates the same concept under a different kind of perturbation, namely the one where the underlying task changes. The prototypical example one should keep in mind in this section is that of transfer learning where a classifier trained on a dataset $p^s(x, y)$ is further trained on a new dataset, say $p^t(x,y)$. We will assume that the input domain of the two distributions is the same.

\subsection{Changing the data distribution}
\label{ss:adapting_data_distribution}

If i.i.d samples from the source task are denoted by $X^s = \cbrac{x^s_1, \ldots, x^s_{n_s}}$ and those of the target distribution are $X^t = \cbrac{x^t_1, \ldots, x^t_{n_t}}$ the empirical source and target distributions can be written as
\[
    p^s(x) = \f{1}{n_s} \sum_{i=1}^{n_s} \delta_{x-x^s_i}, \trm{and}\ p^t(x) = \f{1}{n_t} \sum_{i=1}^{n_t} \delta_{x-x^t_i}
\]
respectively; here $\delta_{x-x'}$ is a Dirac delta distribution at $x'$. We will consider a transport problem that transports the source distribution $p^s(x)$ to the target distribution $p^t(x)$. For any $t \in [0,1]$ we interpolate between the two distributions using a mixture
\beq{
    p(x,t) = (1-t) p^s(x) + t p^t(x).
    \label{eq:interp_p}
}
Observe that the interpolated data distribution equals the source and target distribution at $t=0$ and $t=1$ respectively and it is the mixture of the two distributions for other times. We keep the labels of the data the same and do not interpolate them. As discussed in~\cref{s:app:ot} we can also use techniques from optimal transportation~\citep{villani2008optimal} to obtain a better transport; the same dynamical equations given below remain valid in that case.

\subsection{Iso-classification process with a changing data distribution}
\label{ss:iso_c_data}

The equilibrium surface $\thlg$ in~\cref{fig:rdc} is a function of the task and also evolves with the task. We now give a dynamical process that keeps the model parameters in equilibrium as the task evolves quasi-statically.
We again have the same conditions for the dynamics as those in~\cref{eq:iso_c_conditions}. The following lemma is analogous to~\cref{lem:equilibrium_dynamics}.
\begin{lemma}[Dynamical process for changing data distribution]
\label{lem:dynamics_data}
Given $(\ldot, \gdot)$, the evolution of model parameters $\th$ for a changing data distribution given by~\cref{eq:interp_p} is
\beq{
    \thdot = \th_\l \ldot + \th_\g \gdot + \th_t
    \label{eq:th_dynamics_data}
}
where
\beq{
    \th_t = A^{-1}\ b_t =: -A^{-1} \int \pf[p(x,t)]{t}\ \ag{\grad_\th H}\ \dx
    \label{eq:b_t}
}
and the other quantities are as defined in~\cref{lem:equilibrium_dynamics} with the only change that expectations on data $x$ are taken with respect to $p(x,t)$ instead of $p(x)$. The additional term $\th_t$ arises because the data distribution changes with time.
\end{lemma}

A similar computation as that of~\cref{ss:iso_classification} gives a quasi-static iso-classification process as the task evolves
\beq{
    \aed{
        \thdot &= \th_\l \ldot + \th_\g \gdot + \th_t\\
        0 &= C_\l \ldot + C_\g \gdot + C_t
    }
    \label{eq:iso_c_constraint_data}
}
where $C_\l$ and $C_\g$ are as given in~\cref{eq:cl_cg} with the only change being that the outer expectation is taken with respect to $x \sim p(x,t)$. The new term that depends on time $t$ is
\beq{
    C_t = \scalemath{1}{-
        \int\ \pf[p(x,t)]{t} \ag{\ell} \dx - \E_{x \sim p(x,t)}\ \sqbrac{
        \ag{ \th_t ^{\top}\grad_\th H } \ag{\ell} - \ag{\th_t^{\top}\grad_\th H \ \ell} +
        \ag{\th_t^{\top} \grad_\th \ell}
        }}
    \label{eq:ct}
}
with $\ell = \log c_\th(y_{x_t} | z)$. Finally get
\beq{
    \aed{
        \thdot &= \rbrac{\th_\l - \f{C_\l}{C_\g}\ \th_\g} \ldot + \rbrac{\th_t - \f{C_t}{C_\g}  \th_\g}\\
        &=: \hthl \ldot + \htht
    }.
    \label{eq:thdot_data_combined}
}
This indicates that $\th = \th(\l, t)$ is a surface parameterized by $\l$ and $t$, equipped with a basis of tangent plane $(\hthl, \htht)$.

\subsection{Geodesic transfer of representations}
\label{ss:geodesic_transfer}

The dynamics of~\cref{lem:dynamics_data} is valid for any $(\ldot, \gdot)$. We provide a locally optimal way to change $(\l, \g)$ in this section.

\begin{remark}[Rate-distortion trade-off]
\label{eq:rate_distortion_tradeoff}
Note that
\beq{
    \aed{
        \dot{C} &= 0,\\
        \dot{D} &= \pf[D]{\l} \ldot + \pf[D]{\g} \gdot
        =-\a \rbrac{\ppf[F]{\l} \ppf[F]{\g} - \rbrac{\ppftwo[F]{\partial \l \partial \g}}^2 } =-\a \det \rbrac{\trm{Hess}(F)},\\
        \dot{R} &= \pf[R]{D} \dot{D} + \pf[R]{C} \dot{C} = - \l \dot{D}.
    }
    \label{eq:rdot_isoc}
}
The first equality is simply our iso-classification constraint. For $\a > 0$, the second one indicates that $\dot{D} < 0$ using~\cref{lem:r_cvx_f_noncvx} which shows that $0 \succ \trm{Hess}(F)$. This also gives $\ldot > 0$ in~\cref{eq:dl_dg_isoc}. The third equality is a powerful observation: it indicates a trade-off between rate and distortion, if $\dot{D} < 0$ we have $\dot{R} > 0$. It also shows the geometric structure of the equilibrium surface by connecting $\dot{R}$ and $\dot{D}$ together, which we will exploit next.
\end{remark}

Computing the functionals $R, D$ and $C$ during the iso-classification transfer presents us with a curve in $RDC$ space. Geodesic transfer implies that the functionals $R, D$ follow the shortest path in this space. But notice that if \tbf{we assume that the model capacity is infinite}, the $RDC$ space is Euclidean and therefore the geodesic is simply a straight line. Since we keep the classification loss constant during the transfer, $\dot{C} = 0$, straight line implies that slope $\d D/\d R$ is a constant, say $k$. Thus $\dot{D} = k \dot{R}$. Observe that $\dot{R} = \pf[R]{D} \dot{D} + \pf[R]{C} \dot{C} + \pf[R]{t} = - \l \dot{D} + \pf[R]{t}$. Combining the iso-classification constraint and the fact that $\dot{D} = k \dot{R} = -k \l \dot{D} + k \pf[R]{t}$, gives us a linear system:
\beq{
    \aed{
        &\pf[D]{t} + \pf[D]{\l} \dot{\l} + \pf[D]{\g} \dot{\g} = \frac{ k \pf[R]{t}  }{ 1 + k \l };\\
        & \pf[C]{\l} \dot{\l} + \pf[C]{\g} \dot{\g} + \pf[C]{t} = 0
    }
    \label{eq:geodesic_system}
}
We solve this system to update $(\l, \g)$ during the transfer.

\section{Experimental validation}
\label{s:expt}

\begin{figure*}[!t]
\centering
    \begin{subfigure}[b]{0.32\textwidth}
    \centering
    \includegraphics[width=\textwidth]{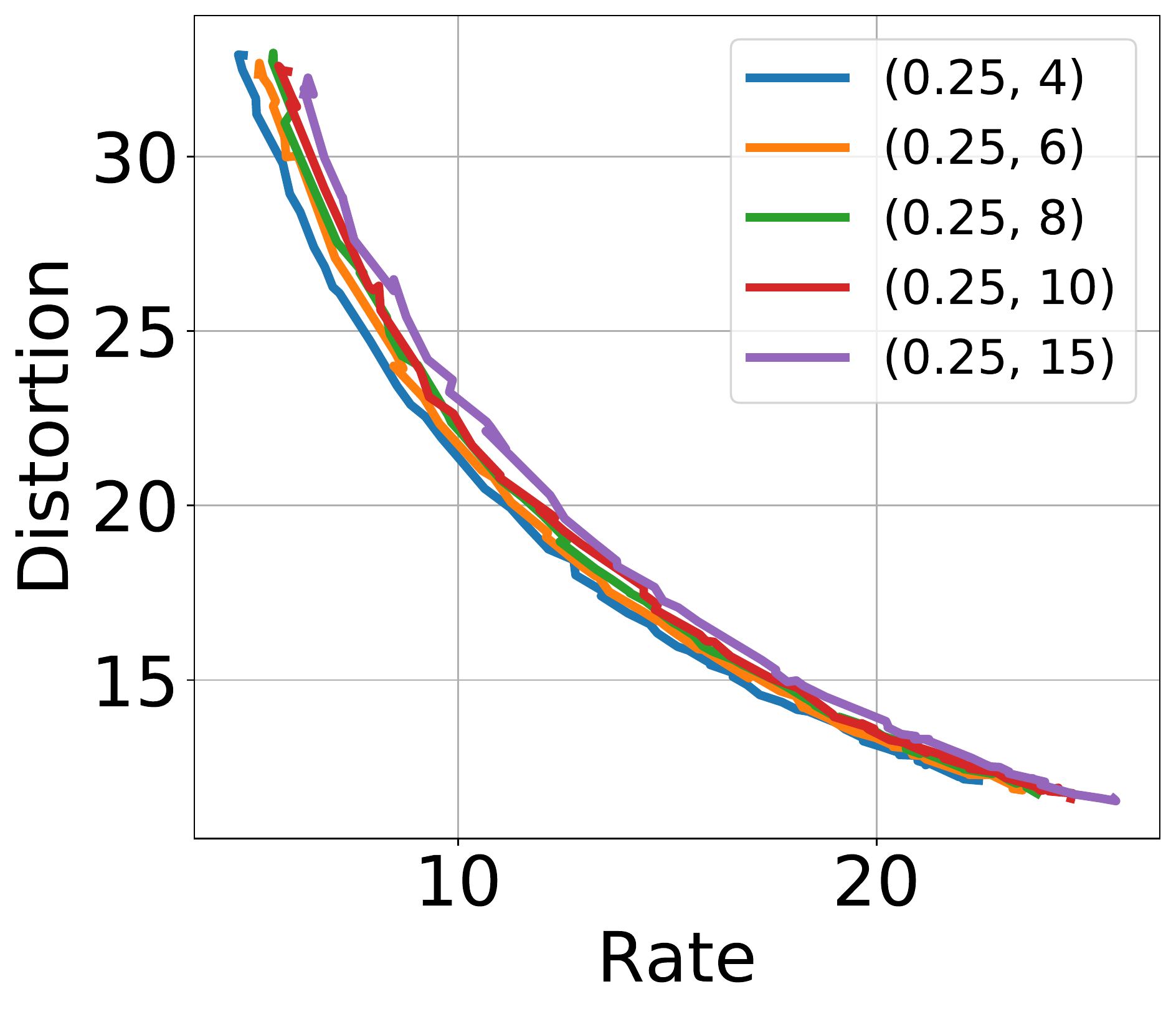}
    \caption{}
    \label{fig:mnist_isoc:rd}
    \end{subfigure}
    \begin{subfigure}[b]{0.32\textwidth}
    \centering
    \includegraphics[width=\textwidth]{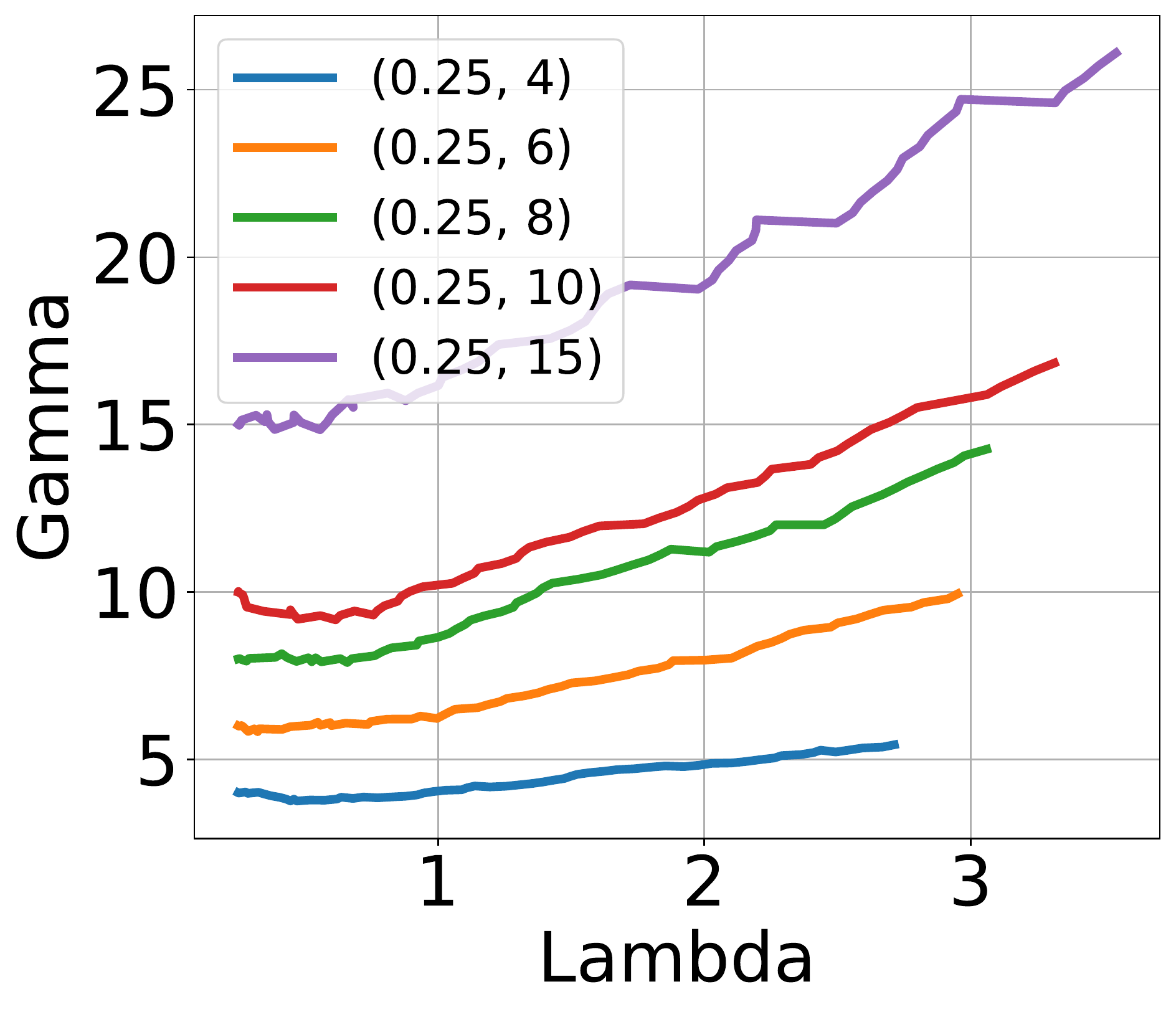}
    \caption{}
    \label{fig:mnist_isoc:lg}
    \end{subfigure}
    \begin{subfigure}[b]{0.33\textwidth}
    \centering
    \includegraphics[width=\textwidth]{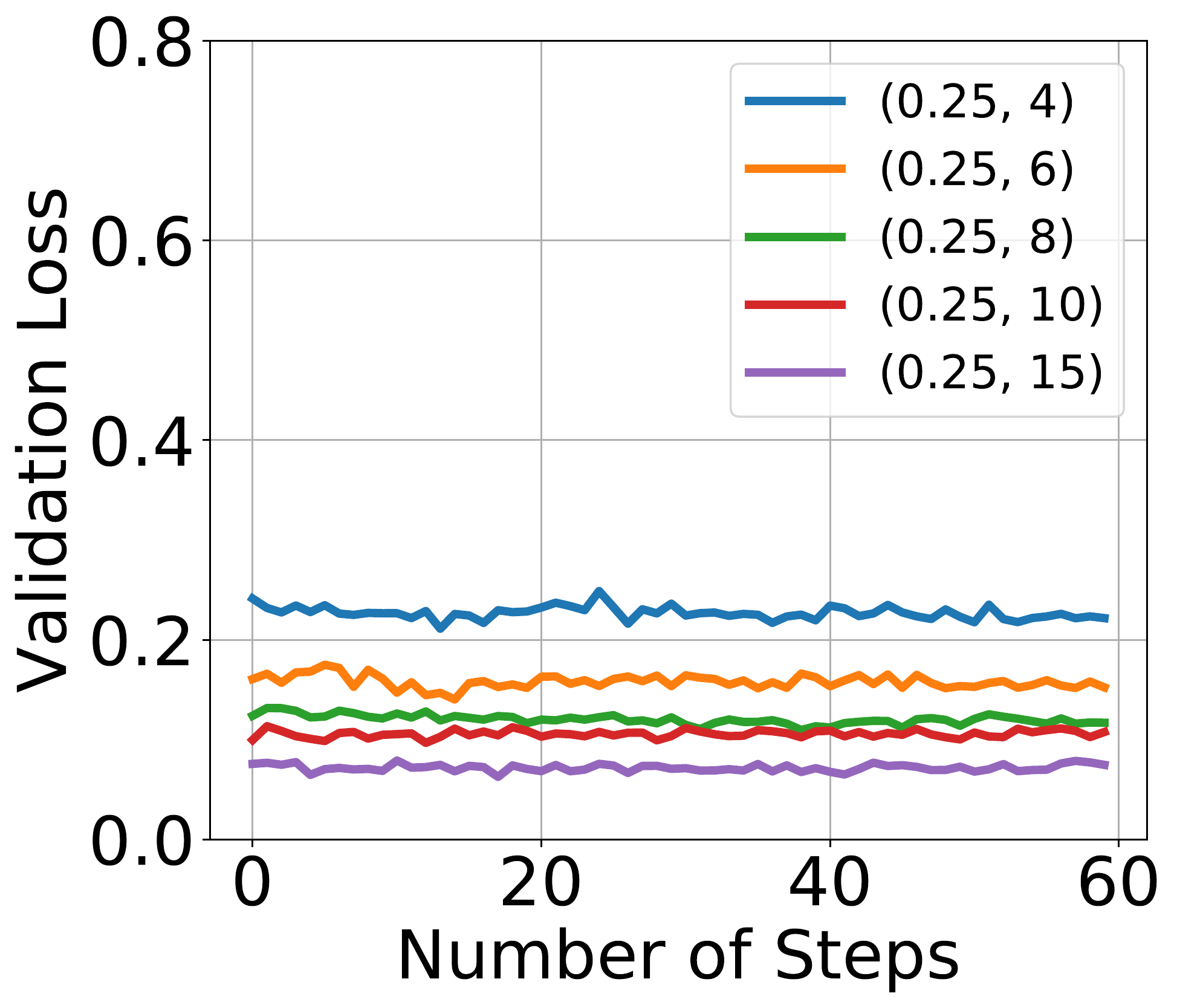}
    \caption{}
    \label{fig:mnist_isoc:c}
    \end{subfigure}
\caption{\tbf{Iso-classification process for MNIST.}
We run 5 different experiments for initial Lagrange multipliers given by $\l=0.25$ and $\g \in \cbrac{4,6,8,10,15}$.
During each experiment, we modify these Lagrange multipliers (\cref{fig:mnist_isoc:lg}) to keep the classification loss constant and plot the rate-distortion curve (\cref{fig:mnist_isoc:rd}) along with the validation loss (\cref{fig:mnist_isoc:c}). The validation accuracy is constant for each experiment; it is between 92--98\% for these initial values of $(\l,\g)$. Similarly the training loss is almost unchanged during each experiment and takes values between 0.06--0.2 for different values of $(\l,\g)$.
}
\label{fig:mnist_isoc}
\end{figure*}

\begin{figure*}[!htpb]
\centering
    \begin{subfigure}[b]{0.32\textwidth}
    \centering
    \includegraphics[width=\textwidth]{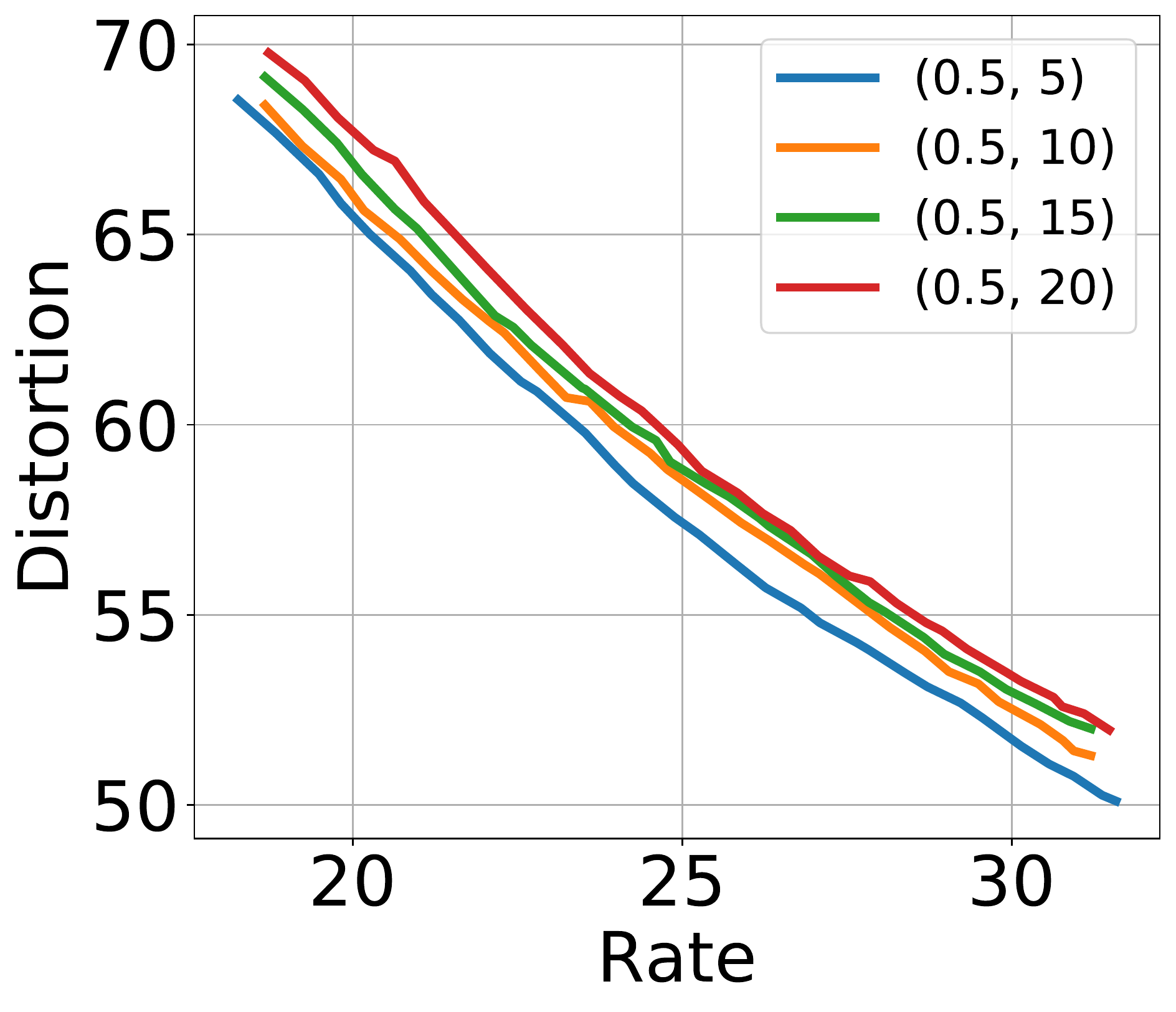}
    \caption{}
    \label{fig:cifar_isoc:rd}
    \end{subfigure}
    \begin{subfigure}[b]{0.32\textwidth}
    \centering
    \includegraphics[width=\textwidth]{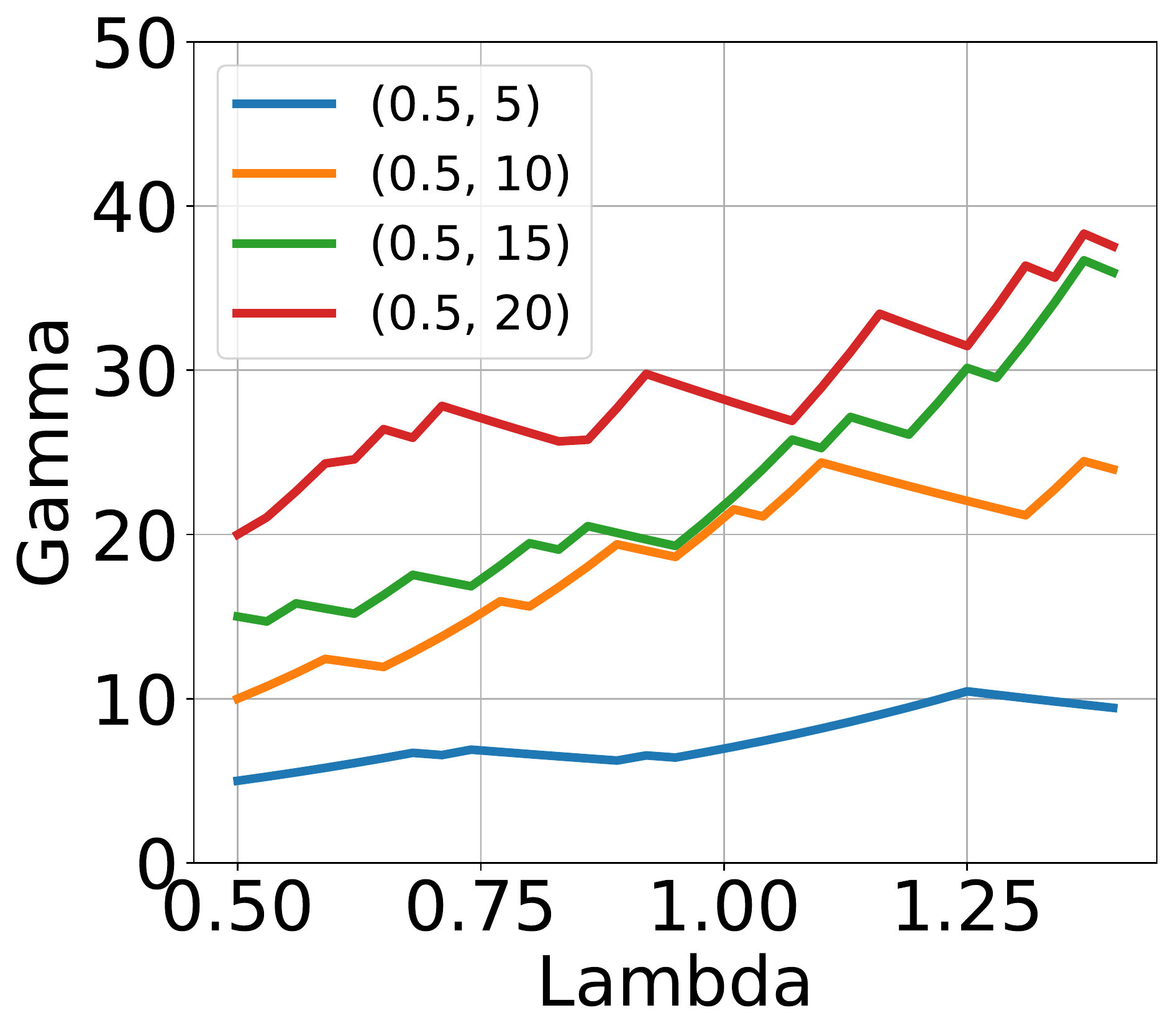}
    \caption{}
    \label{fig:cifar_isoc:lg}
    \end{subfigure}
    \begin{subfigure}[b]{0.33\textwidth}
    \centering
    \includegraphics[width=\textwidth]{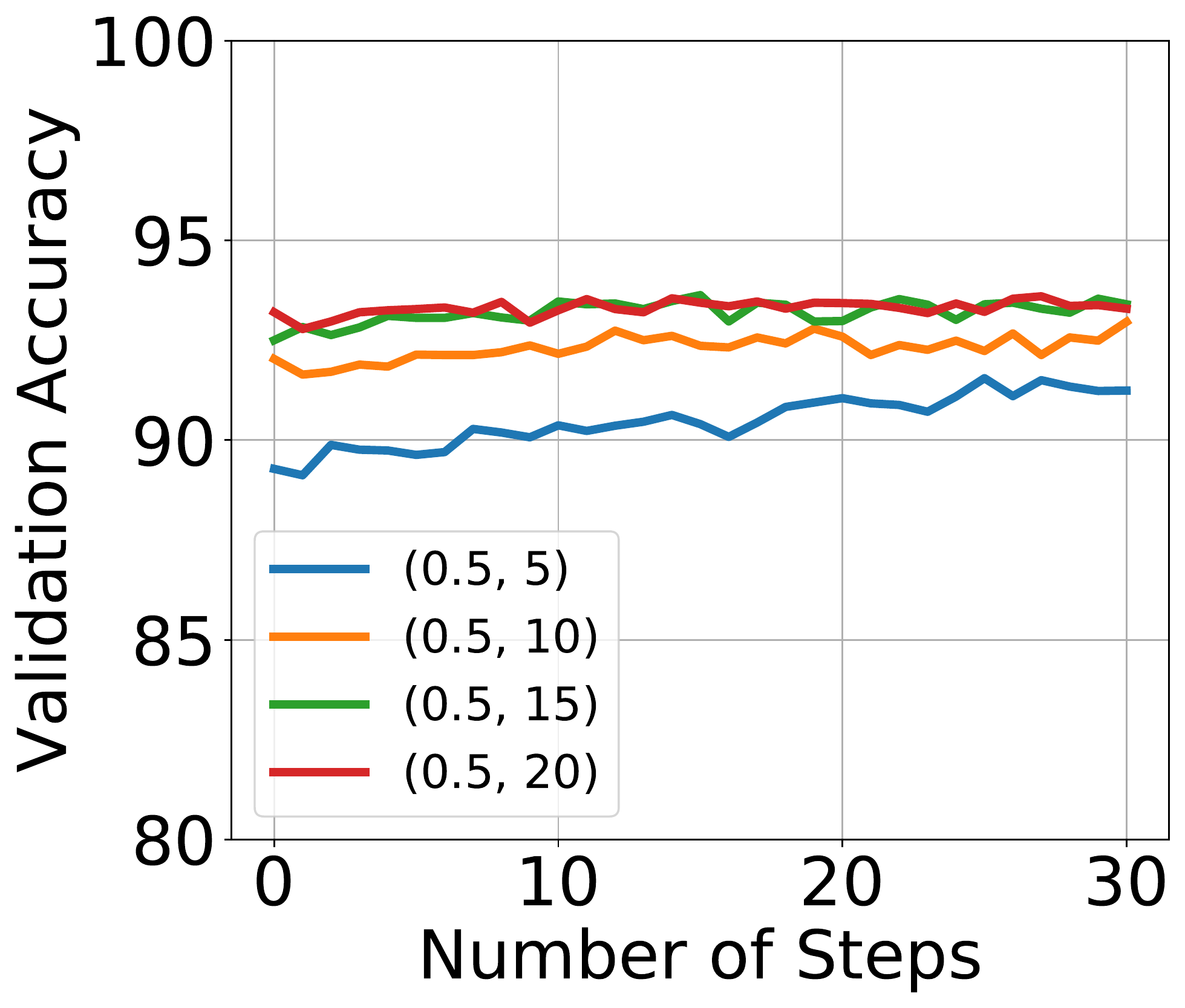}
    \caption{}
    \label{fig:cifar_isoc:acc}
    \end{subfigure}
\caption{\tbf{Iso-classification process for CIFAR-10.} We run 4 different experiments for initial Lagrange multipliers $\l=0.5$ and $\g \in \cbrac{5,10,15,20}$. During each experiment, we modify the Lagrange multipliers (\cref{fig:cifar_isoc:lg}) to keep the classification loss constant and plot the rate-distortion curve (\cref{fig:cifar_isoc:rd}) along with the validation accuracy (\cref{fig:cifar_isoc:acc}). The validation loss is constant during each experiment; it takes values between 0.5--0.8 for these initial values of $(\l,\g)$. Similarly, the training loss is constant and takes values between 0.02--0.09 for these initial values of $(\l,\g)$. Observe that the rate-distortion curve in~\cref{fig:cifar_isoc:rd} is much flatter than the one in~\cref{fig:mnist_isoc:rd} which indicates that the model family $\Th$ for CIFAR-10 is much more powerful; this corresponds to the straight line in the RD curve for an infinite model capacity is as shown in~\cref{fig:rd}.
}
\label{fig:cifar_isoc}
\end{figure*}

This section presents experimental validation for the ideas in this paper. We first implement the dynamics in~\cref{s:dynamical_processes_equilibrium_surface} that traverses the equilibrium surface and then demonstrate the dynamical process for transfer learning devised in~\cref{s:transfer}.

\heading{Setup} We use the MNIST~\citep{lecun1998gradient} and CIFAR-10~\citep{krizhevsky2009learning} datasets for our experiments. We use a 2-layer fully-connected network (same as that of~\citet{kingma2013auto}) as the encoder and decoder for MNIST; the encoder for CIFAR-10 is a ResNet-18~\citep{he2016identity} architecture while the decoder is a 4-layer deconvolutional network~\citep{noh2015learning}. Full details of the pre-processing, network architecture and training are provided in~\cref{s:app:setup}.

\subsection{Iso-classification process on the equilibrium surface}
\label{ss:expt:traveling_on_surface}

\begin{figure}[!tbh]
\centering
    \begin{subfigure}[b]{0.32\textwidth}
    \centering
    \includegraphics[width=\textwidth]{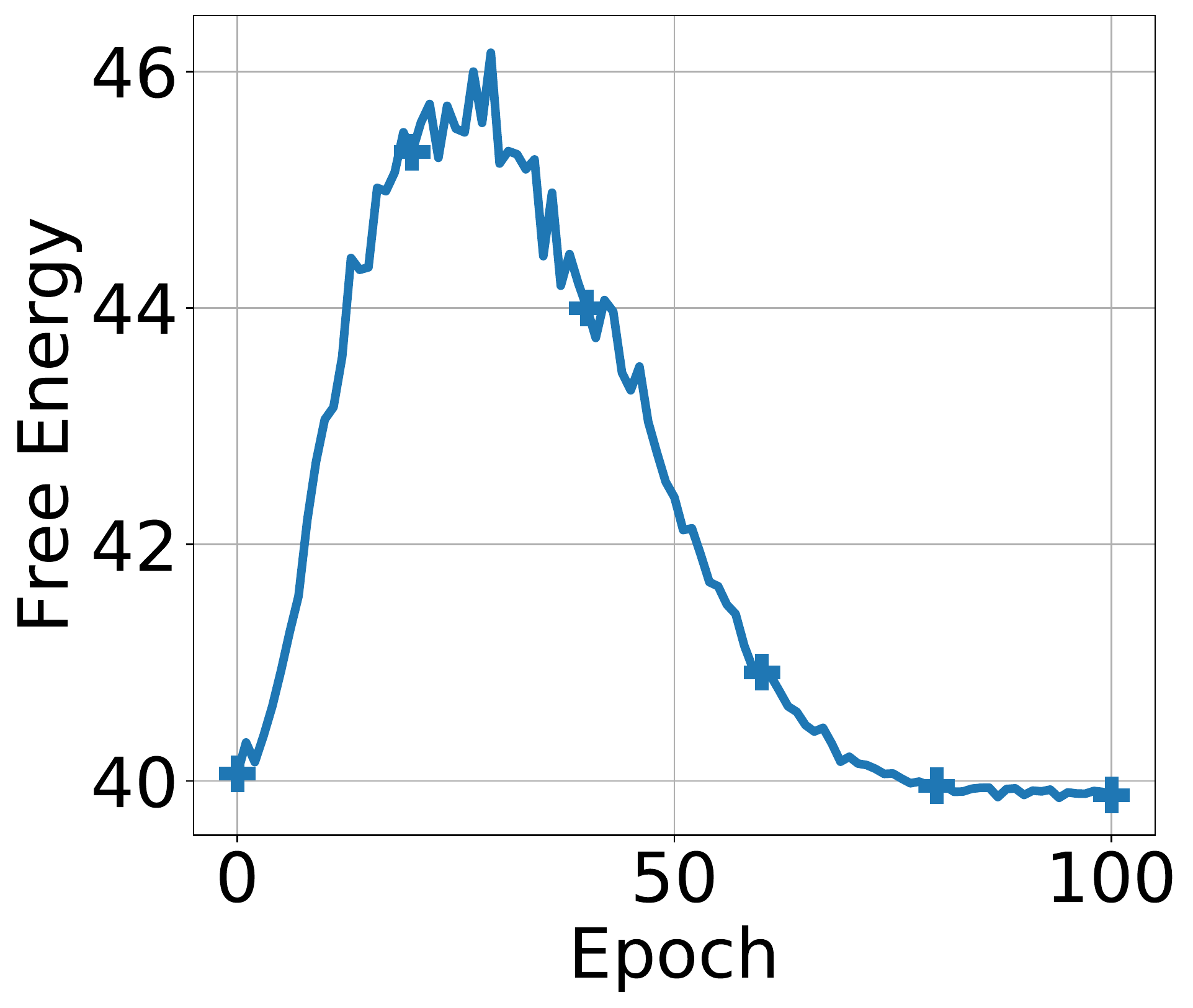}
    \caption{}
    \label{fig:ff_equilibriation}
    \end{subfigure}
    \hspace{0.25in}
    \begin{subfigure}[b]{0.33\textwidth}
    \centering
    \includegraphics[width=\textwidth]{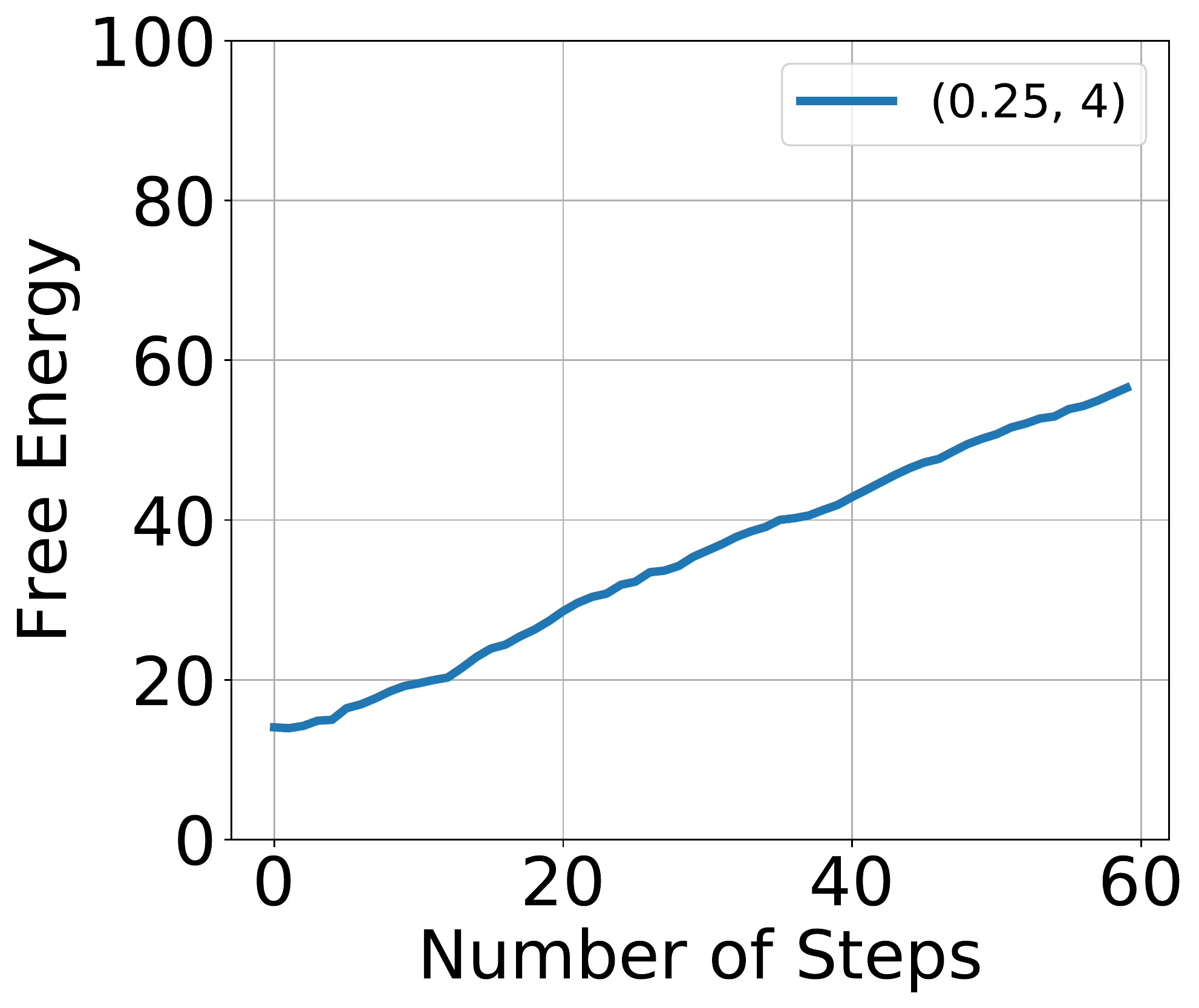}
    \caption{}
    \label{fig:ff_large_change}
    \end{subfigure}
\caption{\tbf{Variation of the free-energy $\ff(\l, \g)$ across the equilibration and the iso-classification processes.} \cref{fig:ff_equilibriation} shows the free-energy during equilibration between small changes of $(\l, \g)$. The initial and final values of the Lagrange multipliers are $(0.5, 1)$ and $(0.51, 1.04)$ respectively and the free-energy is about the same for these values. \cref{fig:ff_large_change} shows the free-energy as $(\l,\g)$ undergo a large change from their initial value of $(0.25,4)$ to $(3.5, 26)$ during the iso-classification process in~\cref{fig:mnist_isoc}. Since the rate-distortion change a lot (\cref{fig:mnist_isoc:rd}), the free-energy also changes a lot even if $C$ is constant (\cref{fig:mnist_isoc:c}). Number of steps in~\cref{fig:ff_large_change} refers to the number of steps of running~\cref{eq:ldot_gdot_isoc}.
}
\label{fig:ff}
\end{figure}

This experiment demonstrates the iso-classification process in~\cref{rem:implementing_iso_c}. As discussed in~\cref{rem:why_equilibrium}, training a model to minimize the functional $R + \l D + \g C$ decreases the free-energy monotonically.

\heading{Details} Given a value of the Lagrange multipliers $(\l,\g)$ we first find a model on the equilibrium surface by training from scratch for 120 epochs with the Adam optimizer~\citep{kingma2014adam}; the learning rate is set to $10^{-3}$ and drops by a factor of 10 every 50 epochs. We then run the iso-classification process for these models in~\cref{rem:implementing_iso_c} as follows. We modify $(\l,\g)$ according to the equations
\beq{
    \ldot = -\a \pf[C]{\g}\quad \trm{and}\quad
    \gdot = \a \pf[C]{\l}.
    \label{eq:ldot_gdot_isoc}
}

Changes in $(\l, \g)$ cause the equilibrium surface to change, so it is necessary to adapt the model parameters $\th$ so as to keep them on the dynamically changing surface; let us call this process of adaptation ``equilibriation''. We achieve this by taking gradient-based updates to minimize $J(\l,\g)$ with a learning rate schedule that looks like a sharp quick increase from zero and then a slow annealing back to zero. The learning rate schedule is given by $\eta(t) = (t/T)^2 \rbrac{1-t/T}^5$ where $t$ is the number of mini-batch updates taken since the last change in $(\l,\g)$ and $T$ is total number of mini-batch updates of equilibration. The maximum value of the learning rate is set to $1.5 \times 10^{-3}$. The free-energy should be unchanged if the model parameters are on the equilibrium surface after equilibration; this is shown in~\cref{fig:ff_equilibriation}. Partial derivatives in~\cref{eq:ldot_gdot_isoc} are computed using finite-differences.

\cref{fig:mnist_isoc} shows the result for the iso-classification process for MNIST and~\cref{fig:cifar_isoc} shows a similar result for CIFAR-10. We can see that the classification loss remains constant through the process. This experiment shows that we can implement an iso-classification process while keeping the model parameters on the equilibrium surface during it.

\subsection{Transferring representations to new data}
\label{ss:expt:transfer}

\begin{figure}[!h]
\centering
    \begin{subfigure}[b]{0.32\textwidth}
    \centering
    \includegraphics[width=\textwidth]{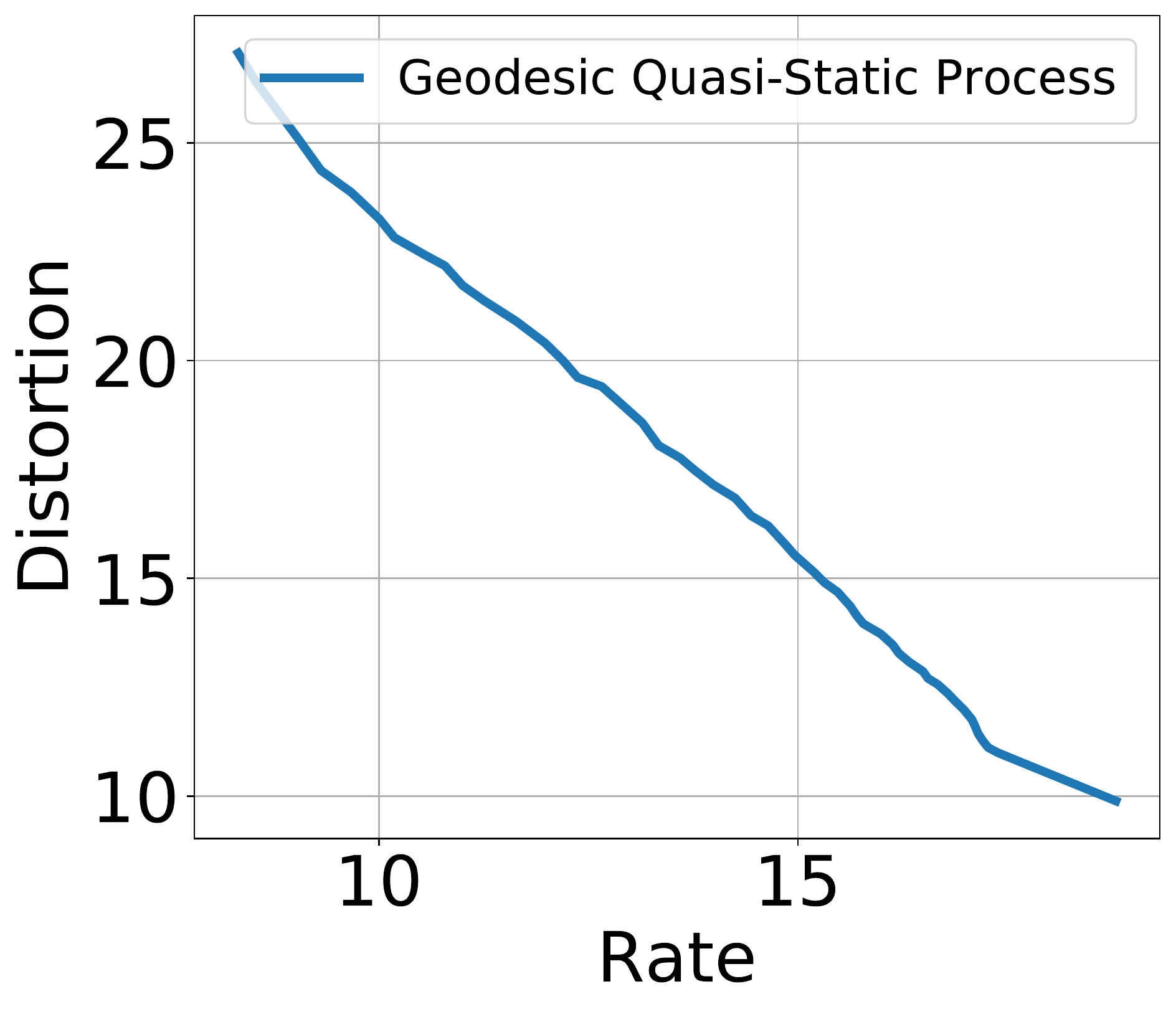}
    \caption{}
    \label{fig:mnist_transfer:rd}
    \end{subfigure}
    \hspace{0.3in}
    \begin{subfigure}[b]{0.33\textwidth}
    \centering
    \includegraphics[width=\textwidth]{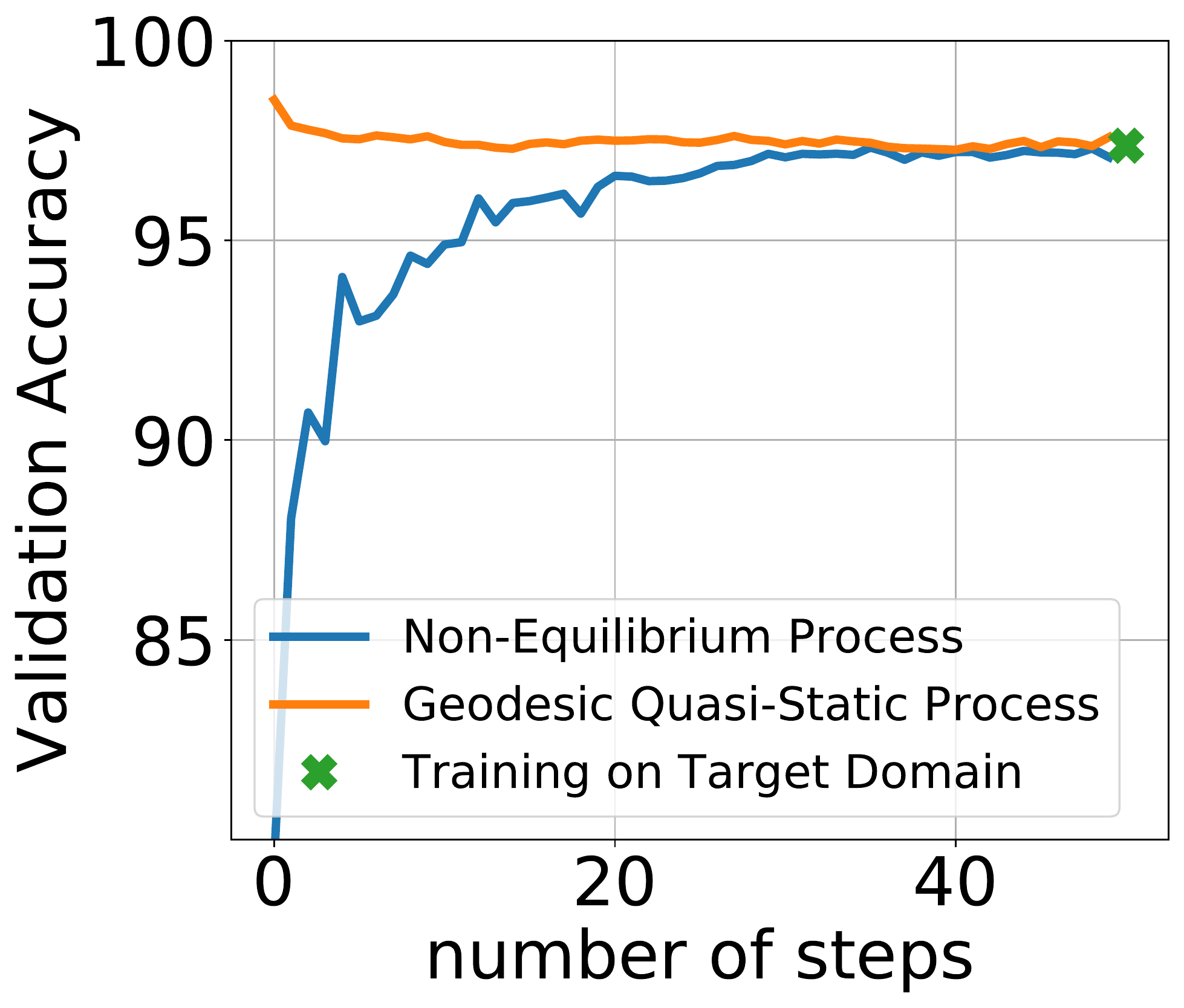}
    \caption{}
    \label{fig:mnist_transfer:acc}
    \end{subfigure}
\caption{\tbf{Transferring from source dataset of MNIST digits 0--4 to the target dataset consisting of digits 5--9.} \cref{fig:mnist_transfer:rd} shows the variation of rate and distortion during the transfer; as discussed in~\cref{ss:geodesic_transfer} we maintain a constant $\d R/\d D$ during the transfer; the rate decreases and the distortion increases. \cref{fig:mnist_transfer:acc} shows the validation accuracy during the transfer. The orange curve corresponds to geodesic iso-classification transfer; the blue curve is the result of directly fine-tuning the source model on the target data (note the very low accuracy at the start); the green point is the accuracy of training on the target task from scratch.}
\label{fig:mnist_transfer}
\end{figure}

We next present experimental results of an iso-classification process for transferring the learnt representation. We pick the source dataset to be all images corresponding to digits 0--4 in MNIST and the target dataset is its complement, images of digits 5--9. Our goal is to adapt a model trained on the source task to the target task while keeping its classification loss constant. We run the geodesic transfer dynamics from~\cref{ss:geodesic_transfer} and the results are shown in~\cref{fig:mnist_transfer}.

It is evident that the classification accuracy is constant throughout the transfer and is also the same as that of training from scratch on the target. MNIST is an simple dataset and the accuracy gap between iso-classification transfer, fine-tuning from the source and training from scratch is minor. The benefit of running the iso-classification transfer however is that we can be guaranteed about the final accuracy of the model. We expect the gap between these three to be significant for more complex datasets.
Results for a similar experiment for transferring between a source dataset that consists of all vehicles in CIFAR-10 to a target dataset that consists of all animals are provided in~\cref{s:app:transfer_cifar_heuristic}.



\section{Related work}
\label{s:related_work}

We are motivated by the Information Bottleneck (IB) principle of~\citet{tishby2000information,shwartz2017opening}, which has been further explored by~\citet{achille2017emergence,alemi2016deep,higgins2016beta}. The key difference in our work is that while these papers seek to understand the representation for a given task, we focus on how the representation can be adapted to a new task. Further, the Lagrangian in~\cref{eq:f_eth_th} has connections to PAC-Bayes bounds~\citep{mcallester2013pac,dziugaite2017computing} and training algorithms that use the free-energy~\citep{chaudhari2019entropy}. Our use of rate-distortion for transfer learning is close to the work on unsupervised learning of~\citet{brekelmans2019exact,ver2015maximally}.

This paper builds upon the work of~\citet{alemi2017fixing,alemi2018therml}. We refine some results therein, viz., we provide a proof of the convexity of the equilibrium surface and identify it with the equilibrium distribution of SGD (\cref{rem:why_equilibrium}). We introduce new ideas such as dynamical processes on the equilibrium surface. Our use of thermodynamics is purely as an inspiration; the work presented here is mathematically rigorous and also provides an immediate algorithmic realization of the ideas.

This paper has strong connections to works that study stochastic processes inspired from statistical physics for machine learning, e.g., approximate Bayesian inference and implicit regularization of SGD~\citep{mandt2017stochastic,chaudhari2017stochastic}, variational inference~\citep{jordan1998introduction,kingma2013auto}. The iso-classification process instantiates an ``automatic'' regularization via the trade-off between rate and distortion; this point-of-view is an exciting prospect for future work.
The technical content of the paper also draws from optimal transportation~\citep{villani2008optimal}.


A large number of applications begin with pre-trained models~\citep{sharif2014cnn,girshick2014rich} or models trained on tasks different~\citep{doersch2017multi}. Current methods in transfer learning however do not come with guarantees over the performance on the target dataset, although there is a rich body of older work~\citep{baxter2000model} and ongoing work that studies this~\citep{zamir2018taskonomy}. The information-theoretic understanding of transfer and the constrained dynamical processes developed in our paper is a first step towards building such guarantees. In this context, our theory can also be used to tackle catastrophic forgetting~\citet{kirkpatrick2017overcoming} to ``detune'' the model post-training and build up redundant features.

\section{Discussion}
\label{s:discussion}

We presented dynamical processes that maintain the parameters of model on an equilibrium surface that arises out of a certain free-energy functional for the encoder-decoder-classifier architecture. The decoder acts as a measure of the information discarded by the encoder-classifier pair while fitting on a given task. We showed how one can develop an iso-classification process that travels on the equilibrium surface while keeping the classification loss constant. We showed an iso-classification transfer learning process which keeps the classification loss constant while adapting the learnt representation from a source task to a target task.

The information-theoretic point-of-view in this paper is rather abstract but its benefit lies in its exploitation of the equilibrium surface. Relationships between the three functionals, namely rate, distortion and classification, that define this surface, as also other functionals that connect to the capacity of the hypothesis class such as the entropy $S$ may allow us to define invariants of the learning process. For complex models such as deep neural networks, such a program may lead an understanding of the principles that govern their working.

{
\footnotesize
\bibliographystyle{apalike}
\bibliography{dl,thermo}

\begin{thebibliography}{}

\bibitem[Achille and Soatto, 2017]{achille2017emergence}
Achille, A. and Soatto, S. (2017).
\newblock On the emergence of invariance and disentangling in deep
  representations.
\newblock {\em arXiv:1706.01350}.

\bibitem[Alemi and Fischer, 2018]{alemi2018therml}
Alemi, A.~A. and Fischer, I. (2018).
\newblock Therml: Thermodynamics of machine learning.
\newblock {\em arXiv preprint arXiv:1807.04162}.

\bibitem[Alemi et~al., 2016]{alemi2016deep}
Alemi, A.~A., Fischer, I., Dillon, J.~V., and Murphy, K. (2016).
\newblock Deep variational information bottleneck.
\newblock {\em arXiv:1612.00410}.

\bibitem[Alemi et~al., 2017]{alemi2017fixing}
Alemi, A.~A., Poole, B., Fischer, I., Dillon, J.~V., Saurous, R.~A., and
  Murphy, K. (2017).
\newblock Fixing a broken elbo.
\newblock {\em arXiv preprint arXiv:1711.00464}.

\bibitem[Baxter, 2000]{baxter2000model}
Baxter, J. (2000).
\newblock A model of inductive bias learning.
\newblock {\em Journal of artificial intelligence research}, 12:149--198.

\bibitem[Brekelmans et~al., 2019]{brekelmans2019exact}
Brekelmans, R., Moyer, D., Galstyan, A., and Ver~Steeg, G. (2019).
\newblock Exact rate-distortion in autoencoders via echo noise.
\newblock In {\em Advances in Neural Information Processing Systems}, pages
  3884--3895.

\bibitem[Chaudhari et~al., 2019]{chaudhari2019entropy}
Chaudhari, P., Choromanska, A., Soatto, S., LeCun, Y., Baldassi, C., Borgs, C.,
  Chayes, J., Sagun, L., and Zecchina, R. (2019).
\newblock Entropy-sgd: Biasing gradient descent into wide valleys.
\newblock {\em Journal of Statistical Mechanics: Theory and Experiment},
  2019(12):124018.

\bibitem[Chaudhari and Soatto, 2017]{chaudhari2017stochastic}
Chaudhari, P. and Soatto, S. (2017).
\newblock Stochastic gradient descent performs variational inference, converges
  to limit cycles for deep networks.
\newblock {\em arXiv preprint arXiv:1710.11029}.

\bibitem[Cuturi, 2013]{cuturi2013sinkhorn}
Cuturi, M. (2013).
\newblock Sinkhorn distances: Lightspeed computation of optimal transport.
\newblock In {\em Advances in neural information processing systems}, pages
  2292--2300.

\bibitem[Doersch and Zisserman, 2017]{doersch2017multi}
Doersch, C. and Zisserman, A. (2017).
\newblock Multi-task self-supervised visual learning.
\newblock In {\em Proceedings of the IEEE International Conference on Computer
  Vision}, pages 2051--2060.

\bibitem[Dziugaite and Roy, 2017]{dziugaite2017computing}
Dziugaite, G.~K. and Roy, D.~M. (2017).
\newblock Computing nonvacuous generalization bounds for deep (stochastic)
  neural networks with many more parameters than training data.
\newblock {\em arXiv preprint arXiv:1703.11008}.

\bibitem[Girshick et~al., 2014]{girshick2014rich}
Girshick, R., Donahue, J., Darrell, T., and Malik, J. (2014).
\newblock Rich feature hierarchies for accurate object detection and semantic
  segmentation.
\newblock In {\em Proceedings of the IEEE conference on computer vision and
  pattern recognition}, pages 580--587.

\bibitem[He et~al., 2016]{he2016identity}
He, K., Zhang, X., Ren, S., and Sun, J. (2016).
\newblock Identity mappings in deep residual networks.
\newblock {\em arXiv:1603.05027}.

\bibitem[Higgins et~al., 2017]{higgins2016beta}
Higgins, I., Matthey, L., Pal, A., Burgess, C., Glorot, X., Botvinick, M.,
  Mohamed, S., and A, L. (2017).
\newblock {beta-VAE: Learning Basic Visual Concepts with a Constrained
  Variational Framework }.
\newblock In {\em ICLR}.

\bibitem[Ioffe and Szegedy, 2015]{ioffe2015batch}
Ioffe, S. and Szegedy, C. (2015).
\newblock {Batch normalization: Accelerating deep network training by reducing
  internal covariate shift}.
\newblock {\em arXiv:1502.03167}.

\bibitem[Jordan et~al., 1998]{jordan1998introduction}
Jordan, M.~I., Ghahramani, Z., Jaakkola, T.~S., and Saul, L.~K. (1998).
\newblock An introduction to variational methods for graphical models.
\newblock In {\em Learning in graphical models}, pages 105--161. Springer.

\bibitem[Kingma and Ba, 2014]{kingma2014adam}
Kingma, D. and Ba, J. (2014).
\newblock Adam: A method for stochastic optimization.
\newblock {\em arXiv:1412.6980}.

\bibitem[Kingma and Welling, 2013]{kingma2013auto}
Kingma, D.~P. and Welling, M. (2013).
\newblock {Auto-encoding variational Bayes}.
\newblock {\em arXiv:1312.6114}.

\bibitem[Kirkpatrick et~al., 2017]{kirkpatrick2017overcoming}
Kirkpatrick, J., Pascanu, R., Rabinowitz, N., Veness, J., Desjardins, G., Rusu,
  A.~A., Milan, K., Quan, J., Ramalho, T., Grabska-Barwinska, A., et~al.
  (2017).
\newblock Overcoming catastrophic forgetting in neural networks.
\newblock {\em Proceedings of the national academy of sciences},
  114(13):3521--3526.

\bibitem[Krizhevsky, 2009]{krizhevsky2009learning}
Krizhevsky, A. (2009).
\newblock Learning multiple layers of features from tiny images.
\newblock Master's thesis, Computer Science, University of Toronto.

\bibitem[LeCun et~al., 1998]{lecun1998gradient}
LeCun, Y., Bottou, L., Bengio, Y., and Haffner, P. (1998).
\newblock Gradient-based learning applied to document recognition.
\newblock {\em Proceedings of the IEEE}, 86(11):2278--2324.

\bibitem[Levin and Peres, 2017]{levin2017markov}
Levin, D.~A. and Peres, Y. (2017).
\newblock {\em Markov chains and mixing times}, volume 107.
\newblock American Mathematical Soc.

\bibitem[Mandt et~al., 2017]{mandt2017stochastic}
Mandt, S., Hoffman, M.~D., and Blei, D.~M. (2017).
\newblock {Stochastic Gradient Descent as Approximate Bayesian Inference}.
\newblock {\em arXiv:1704.04289}.

\bibitem[McAllester, 2013]{mcallester2013pac}
McAllester, D. (2013).
\newblock A pac-bayesian tutorial with a dropout bound.
\newblock {\em arXiv:1307.2118}.

\bibitem[Mezard and Montanari, 2009]{mezard2009information}
Mezard, M. and Montanari, A. (2009).
\newblock {\em Information, physics, and computation}.
\newblock Oxford University Press.

\bibitem[Noh et~al., 2015]{noh2015learning}
Noh, H., Hong, S., and Han, B. (2015).
\newblock Learning deconvolution network for semantic segmentation.
\newblock In {\em Proceedings of the IEEE international conference on computer
  vision}, pages 1520--1528.

\bibitem[Pearlmutter, 1994]{pearlmutter1994fast}
Pearlmutter, B.~A. (1994).
\newblock Fast exact multiplication by the hessian.
\newblock {\em Neural computation}, 6(1):147--160.

\bibitem[Robbins and Monro, 1951]{robbins1951stochastic}
Robbins, H. and Monro, S. (1951).
\newblock A stochastic approximation method.
\newblock {\em The annals of mathematical statistics}, pages 400--407.

\bibitem[Sethna, 2006]{sethna2006statistical}
Sethna, J. (2006).
\newblock {\em Statistical mechanics: entropy, order parameters, and
  complexity}, volume~14.
\newblock Oxford University Press.

\bibitem[Sharif~Razavian et~al., 2014]{sharif2014cnn}
Sharif~Razavian, A., Azizpour, H., Sullivan, J., and Carlsson, S. (2014).
\newblock Cnn features off-the-shelf: an astounding baseline for recognition.
\newblock In {\em Proceedings of the IEEE conference on computer vision and
  pattern recognition workshops}, pages 806--813.

\bibitem[Shwartz-Ziv and Tishby, 2017]{shwartz2017opening}
Shwartz-Ziv, R. and Tishby, N. (2017).
\newblock Opening the black box of deep neural networks via information.
\newblock {\em arXiv:1703.00810}.

\bibitem[Tishby et~al., 2000]{tishby2000information}
Tishby, N., Pereira, F.~C., and Bialek, W. (2000).
\newblock The information bottleneck method.
\newblock {\em arXiv preprint physics/0004057}.

\bibitem[Ver~Steeg and Galstyan, 2015]{ver2015maximally}
Ver~Steeg, G. and Galstyan, A. (2015).
\newblock Maximally informative hierarchical representations of
  high-dimensional data.
\newblock In {\em Artificial Intelligence and Statistics}, pages 1004--1012.

\bibitem[Villani, 2008]{villani2008optimal}
Villani, C. (2008).
\newblock {\em Optimal transport: old and new}, volume 338.
\newblock Springer Science \& Business Media.

\bibitem[Zamir et~al., 2018]{zamir2018taskonomy}
Zamir, A.~R., Sax, A., Shen, W., Guibas, L.~J., Malik, J., and Savarese, S.
  (2018).
\newblock Taskonomy: Disentangling task transfer learning.
\newblock In {\em Proceedings of the IEEE Conference on Computer Vision and
  Pattern Recognition}, pages 3712--3722.

\end{thebibliography}
}
\clearpage
\begin{appendix}
\onecolumn

\begin{center}
\vspace*{0.5in}
\begin{Large}
\tbf{Appendix}
\end{Large}
\vspace*{0.5in}
\end{center}

\section{Details of the experimental setup}
\label{s:app:setup}

\tbf{Datasets.} We use the MNIST~\citep{lecun1998gradient} and CIFAR-10~\citep{krizhevsky2009learning} datasets for these experiments. The former consists of 28 $\times$28-sized gray-scale images of handwritten digits (60,000 training and 10,000 validation). The latter consists of 32$\times$32-sized RGB images (50,000 training and 10,000 for validation) spread across 10 classes; 4 of these classes (airplane, automobile, ship, truck) are transportation-based while the others are images of animals and birds.

\tbf{Architecture and training.} All models in our experiments consist of an encoder-decoder pair along with a classifier that takes in the latent representation as input. For experiments on MNIST, both encoder and decoder are multi-layer perceptrons with 2 fully-connected layers, the decoder uses a mean-square error loss, i.e., a Gaussian reconstruction likelihood and the classifier consists of a single fully-connected layer. For experiments on CIFAR-10, we use a residual network~\citep{he2016identity} with 18 layers as an encoder and a decoder with one fully-connected layer and 4 deconvolutional layers~\citep{noh2015learning}. The classifier network for CIFAR-10 is a single fully-connected layer. All models use ReLU non-linearities and batch-normalization~\citep{ioffe2015batch}. Further details of the architecture are given in~\cref{s:app:setup}. We use Adam~\citep{kingma2014adam} to train all models with cosine learning rate annealing.

The encoder and decoder for MNIST has 784--256--16 neurons on each layer; the encoding $z$ is thus 16-dimensional which is the input to the decoder. The classifier has one hidden layer with 12 neurons and 10 outputs. The encoder for CIFAR-10 is a 18-layer residual neural network (ResNet-18) and the decoder has 4 deconvolutional layers. We used a slightly larger network for the geodesic transfer learning experiment on MNIST. The encoder and decoder have 784--400--64 neurons in each layer with a dropout of probability 0.1 after the hidden layer. The classifier has a single layer that takes the 64-dimensional encoding and predicts 10 classes.

\section{Proof of~\cref{lem:r_cvx_f_noncvx}}
\label{s:app:proof_r_cvx_f_noncvx}
The second statement directly follows by observing that $F$ is a minimum of affine functions in $(\l, \g)$. To see the first, evaluate the Hessian of $R$ and $F$
\[
    \aed{
        \trm{Hess}(R)\ \trm{Hess}(F) &=
        \pmat{\ppf[R]{D} & \ppftwo[R]{\partial D \partial C}\\[0.05in] \ppftwo[R]{\partial C \partial D} &\ppf[R]{C}}\
        \pmat{\ppf[F]{\l} & \ppftwo[F]{\partial \l \partial \g}\\[0.05in]
        \ppftwo[F]{\partial {\g} \partial\l} & \ppf[F]{\g}}
    }
\]
Since we have $F = \min_{\eeth, \ddth, \mmth} R + \l D + \g C$, we obtain
\[
    \l = -\pf[R]{D}, \quad \g = -\pf[R]{C}, \quad D = \pf[F]{\l}, \quad C = \pf[F]{\g}.
\]
We then have
\[
    \aed{
        \d \l = -\d \rbrac{\pf[R]{D}}
        &= - \ppf[R]{D}\  \d D - \ppftwo[R]{\partial D \partial C}\ \d C\\
        & = - \ppf[R]{D}\ \rbrac{ \pf[D]{\l}d \l + \pf[D]{\g}d \g }
            - \ppftwo[R]{\partial D \partial C}\ \rbrac{ \pf[C]{\l}d \l + \pf[C]{\g}d \g  } \\
        &= - \rbrac{\ppf[R]{D} \ppf[F]{\l} + \ppftwo[R]{\partial D \partial C} \ppftwo[F]{\partial \g \partial \l}  }\ \d \l
        - \rbrac{ \ppf[R]{D}  \ppftwo[F]{\partial \l \partial \g} + \ppftwo[R]{\partial D \partial C} \ppf[F]{\g}}\ \d \g;
    }
\]
\[
    \aed{
        \d \g = -\d \rbrac{\pf[R]{C}}
        &= - \ppftwo[R]{\partial C \partial D}\ \d D - \ppf[R]{C}\ \d C\\
        & = - \ppftwo[R]{\partial C \partial D}\ \rbrac{ \pf[D]{\l}d \l + \pf[D]{\g}d \g }
            - \ppf[R]{C}\ \rbrac{ \pf[C]{\l}d \l + \pf[C]{\g}d \g  } \\
        &= - \rbrac{\ppftwo[R]{\partial C \partial D} \ppf[F]{\l} + \ppf[R]{C} \ppftwo[F]{\partial \g \partial \l}  }\ \d \l
        - \rbrac{ \ppftwo[R]{\partial C \partial D}  \ppftwo[F]{\partial \l \partial \g} + \ppf[R]{  C} \ppf[F]{\g}}\ \d \g.
    }
\]
Compare the coefficients on both sides to get
\[
  \aed{
      &\ppf[R]{D} \ppf[F]{\l} + \ppftwo[R]{\partial D \partial C} \ppftwo[F]{\partial \g \partial \l} = \ppftwo[R]{\partial C \partial D}  \ppftwo[F]{\partial \l \partial \g} + \ppf[R]{  C} \ppf[F]{\g} = -1; \\
      & \ppf[R]{D}  \ppftwo[F]{\partial \l \partial \g} + \ppftwo[R]{\partial D \partial C} \ppf[F]{\g} = \ppftwo[R]{\partial C \partial D} \ppf[F]{\l} + \ppf[R]{C} \ppftwo[F]{\partial \g \partial \l} = 0,
   }
\]
therefore
\[
    \trm{Hess}(R)\ \trm{Hess}(F) = -I.
\]
Since $0 \succ \trm{Hess}(F)$, we have that $\trm{Hess}(R) \succ 0$, then the constraint surface $f(R, D, C) = 0$ is convex.

\section{Proof of~\cref{lem:equilibrium_dynamics}}
\label{s:app:proof_equilibrium_dynamics}
Recall the definition of the  objective function~\cref{eq:j}, first we compute the gradient of the objective function as following:
\[
\aed{
     \grad_\th \jthlg &= - \E_{x \sim p(x)} \grad_\th \log \zthx \\
     & = - \E_{x \sim p(x)}\f{1}{\zthx} \grad_\th  \zthx \\
     & = -  \E_{x \sim p(x)}  \f{1}{\zthx}\   \int\ ( - \grad_\th H)\ \exp(-H)\ \dz  \\
     & = \E_{x \sim p(x)}  \ag{\grad_\th H}
    \label{eq:grad_j}
}
\]
Then with some effort of computation, we get
\[
\aed{
A = \grad_\th^2 \jthlg &=\grad_\th \E_{x \sim p(x)} \Sqbrac{ \f{1}{\zthx}  \int\ \grad_\th H\ \exp(-H)\ \dz} \\
&= \scalemath{0.6}{\E_{x \sim p(x)}
\sqbrac{ - \f{1}{ \zthx ^2}\ \left(\int\ (- \grad_\th H)\ \exp(-H)\ \dz\right) \left(\int\ \grad_\th^T H\ \exp(-H)\ \dz\right) + \f{1}{ \zthx }\ \int\  \grad_\th^2 H\ \exp(-H)\ \dz - \f{1}{ \zthx }\ \int\ \grad_\th H\ \grad_\th^{\top} H\  \exp(-H)\ \dz}}
 \\
& = \E_{x \sim p(x)} \sqbrac{ \ag{\grad_\th^2 H} + \ag{\grad_\th H} \ag{\grad_\th H}^\top - \ag{\grad_\th H\ \grad_\th^\top H}};\\
b_\l = - \pf{\l} \grad_\th J &=-\pf{\l}
\E_{x \sim p(x)} \sqbrac{ \f{1}{\zthx}\  \int\ \grad_\th H\ \exp(-H)\ \dz} \\
&=- \scalemath{0.6}{\E_{x \sim p(x)}
\sqbrac{ - \f{1}{ \zthx ^2}\ \left(\int\ - \pf[H]{\l}\ \exp(-H)\ \dz\right) \left(\int\ \grad_\th H\ \exp(-H)\ \dz\right) + \f{1}{ \zthx }\ \int\  \pf{\l}\grad_\th H\ \exp(-H)\ \dz - \f{1}{ \zthx }\ \int\ \pf[H]{\l}\ \grad_\th H\  \exp(-H)\ \dz}}
 \\
&= {-\E_{x \sim p(x)} \sqbrac{ \ag{\pf[\grad_\th H]{\l}} - \ag{\pf[H]{\l}\ \grad_\th H} + \ag{\pf[H]{\l}} \ag{\grad_\th H} }};\\
b_\g = - \pf{\g} \grad_\th J &=-\pf{\g}
\E_{x \sim p(x)} \Sqbrac{ \f{1}{\zthx}\  \int\ \grad_\th H\ \exp(-H)\ \dz} \\
&= -\scalemath{0.6}{\E_{x \sim p(x)}
\sqbrac{ - \f{1}{ \zthx ^2}\ \left(\int\ - \pf[H]{\g}\  \exp(-H)\ \dz\right) \left(\int\ \grad_\th H\ \exp(-H)\ \dz\right) + \f{1}{ \zthx }\ \int\  \pf{\g}\ \grad_\th H\ \exp(-H)\ \dz - \f{1}{ \zthx }\ \int\ \pf[H]{\g}\ \grad_\th H\  \exp(-H)\ \dz}}
 \\
&= {-\E_{x \sim p(x)} \sqbrac{ \ag{\pf[\grad_\th H]{\g}} - \ag{\pf[H]{\g}\ \grad_\th H} + \ag{\pf[H]{\g}} \ag{\grad_\th H} }}.
}
\]
According to the quasi-static constraints~\cref{eq:quasi_static_pde}, we have
\[
\aed{
A \thdot - \ldot b_\l - \gdot b_\g = 0,
}
\]
that implies
\beq{
    \aed{
        \thdot &= A^{-1} b_\l\ \ldot + A^{-1} b_\g\ \gdot = \th_\l \ldot + \th_\g \gdot.
    }
}
\section{Computation of Iso-classification constraint}
\label{s:app:iso_c_constraint}
We start with computing the gradient of classification loss, clear that $C = \E_{x \sim p(x)}\sqbrac{-\int\ \dz\ \ee \log \cc} = - \E_{x \sim p(x)}\ag{ \ell }$, where $\ell = \log \ccxth$ is the logarithm of the classification loss, then

\[
    \aed{
        \grad_\th C &=- \grad_\th \E_{x \sim p(x)} \sqbrac{ \f{1}{\zthx}\  \int\ \ell\ \exp(-H)\ \dz} \\
        &= - \scalemath{0.8}{\E_{x \sim p(x)}
        \sqbrac{ - \f{1}{ \zthx ^2}\ \left(\int\ (- \grad_\th H)\ \exp(-H)\ \dz\right) \left(\int\ \ell\ \exp(-H)\ \dz\right) + \f{1}{ \zthx }\ \int\  \grad_\th\ \ell\ \exp(-H)\ \dz - \f{1}{ \zthx }\ \int\ \ell\ \grad_\th H\   \exp(-H)\ \dz}}
         \\
        & = - \E_{x \sim p(x)} \sqbrac{ \ag{\grad_\th \ \ell} + \ag{\grad_\th H} \ag{\ell} - \ag{\ell\ \grad_\th H}};\\
        \pf{\l} C &= - \pf{\l}
        \E_{x \sim p(x)} \sqbrac{ \f{1}{\zthx}\  \int\ \ell\ \exp(-H)\ \dz} \\
        &=- \scalemath{0.8}{\E_{x \sim p(x)}
        \sqbrac{ - \f{1}{ \zthx ^2}\ \left(\int\ - \pf[H]{\l}\ \exp(-H)\ \dz\right) \left(\int\ \ell \ \exp(-H)\ \dz\right)  - \f{1}{ \zthx }\ \int\ \ell\ \pf[H]{\l}\  \exp(-H)\ \dz}}
         \\
        &= -{\E_{x \sim p(x)} \sqbrac{   \ag{\pf[H]{\l}} \ag{ \ell } - \ag{ \ell\ \pf[H]{\l}} }};
    }
\]

\[
    \aed{
        \pf{\g}  C &= - \pf{\g}
        \E_{x \sim p(x)} \sqbrac{ \f{1}{\zthx}\  \int\ \ell\ \exp(-H)\ \dz} \\
        &=- \scalemath{1}{\E_{x \sim p(x)}
        \sqbrac{ - \f{1}{ \zthx ^2}\ \left(\int\ - \pf[H]{\l}\ \exp(-H)\ \dz\right) \left(\int\ \ell \ \exp(-H)\ \dz\right)  - \f{1}{ \zthx }\ \int\ \ell\ \pf[H]{\g}\  \exp(-H)\ \dz}}
         \\
        &= -{\E_{x \sim p(x)} \sqbrac{   \ag{\pf[H]{\g}} \ag{ \ell } - \ag{ \ell\ \pf[H]{\g}} }
    }.
}
\]

The iso-classification loss constrains together with quasi-static constrains imply that:
\[
\aed{
 0 & \equiv \df{t} C \\
 &= \thdot^{\top}\ \grad_\th C + \ldot \pf[C]{ \l} + \gdot \pf[C]{ \g} \\
 & = \ldot \left( \th_\l^{\top}\ \grad_\th C +  \pf[C]{ \l} \right) + \gdot \left( \th_\g^{\top}\ \grad_\th C +  \pf[C]{ \g} \right)
  \\
& = \scalemath{0.68}{
            - \ldot\E_{x \sim p(x)} \sqbrac{
            \ag{\pf[H]{\l}} \ag{\ell} - \ag{ \ell\ \pf[H]{\l}} + \ag{\th_\l^{\top}  \grad_\th H} \ag{\ell} - \ag{ \ell \th_\l^{\top} \grad_\th H } + \ag{\th_\l^{\top} \grad_\th \ell} }}
            \scalemath{0.68}{
            - \gdot\E_{x \sim p(x)} \sqbrac{
            \ag{\pf[H]{\g}} \ag{\ell} - \ag{ \ell\ \pf[H]{\g}} + \ag{\th_\g^{\top}  \grad_\th H} \ag{\ell} - \ag{ \ell \th_\g^{\top} \grad_\th H } + \ag{\th_\g^{\top} \grad_\th \ell} }}\\
& = C_\l \ldot + C_\g \gdot,
}
\]
where the third equation is followed by the equilibrium dynamics~\cref{eq:th_equilibrium_dynamics} for parameters $\th$. So far we developed the constrained dynamics for iso-classification process:
\beq{
    \aed{
        0 &= C_\l \ldot + C_\g \gdot\\
        \thdot &= \th_\l \ldot + \th_\g \gdot.
    }
    \label{eq:iso_c_constraint}
}

\section{Iso-classification equations for changing data distribution}
\label{s:app:iso_c_data}
In this section we analyze the dynamics for iso-classification loss process when the data distribution evolves with time. $\pf[p(x)]{t}$ will lead to additional terms that represent the partial derivatives with respect to $t$ on both the quasi-static and iso-classification constrains. More precisely, the new terms are
\[
\aed{
b_t &=  -\pf{t} \grad_\th J  = - \int\ \pf[p(x)]{t } \ag{ \grad_\th H } \dx; \\
\pf{t}C &= - \int\ \pf[p(x)]{t } \ag{ \ell } \dx,
}
\]
then the quasi-static and iso-classification constraints are ready to be modified as
\[
  \aed{
      0 \equiv \df{t} \grad_\th \jthlg &\Longleftrightarrow 0 = \grad_\th^2 F\ \thdot + \ldot\ \pf[\grad_\th F]{ \l} + \gdot\ \pf[\grad_\th F]{ \g} + \pf[\grad_\th F]{ t} \\
      & \Longleftrightarrow  \thdot =  \ldot\ A^{-1}\ b_\l  + \gdot\ A^{-1}\ b_\g + A^{-1}\ b_t \\
      & \Longleftrightarrow \thdot = \ldot\ \th_\l + \gdot\ \th_\g + \th_t;\\
      0 \equiv \df{t} C &\Longleftrightarrow 0 = \thdot^{\top}\ \grad_\th C + \ldot \pf[C]{ \l} + \gdot \pf[C]{ \g} + \pf[C]{ t} \\
      & \Longleftrightarrow  0 = \ldot \left( \th_\l^{\top}\ \grad_\th C  +  \pf[C]{ \l}\right) + \gdot \left( \th_\g^{\top}\ \grad_\th C  +  \pf[C]{ \g}\right) + \left( \th_t^{\top}\ \grad_\th C +  \pf[C]{ t} \right) \\
      & \Longleftrightarrow  0 = \ldot\ C_\l + \gdot\ C_\g + C_t,
      }
\]
where $A$, $b_\l$, $b_\g$, $C_\l$ and $C_\g$
where $C_\l$ and $C_\g$ are as given in lemma \ref{lem:equilibrium_dynamics} and ~\cref{eq:cl_cg} with the only change being that the outer expectation is taken with respect to $x \sim p(x,t)$. The new terms that depends on time $t$ are
\beq{
    C_t = \scalemath{1}{-
        \int\ \pf[p(x,t)]{t} \ag{\ell} \dx - \E_{x \sim p(x,t)}\ \sqbrac{
        \ag{ \th_t ^{\top}\grad_\th H } \ag{\ell} - \ag{\th_t^{\top}\grad_\th H \ \ell} +
        \ag{\th_t^{\top} \grad_\th \ell}
        }}
    \label{eq:ct}
}
with $\ell = \log c_\th(y_{x_t} | z)$. We can combine modified quasi-static and iso-classification constraints to get
\beq{
    \aed{
        \thdot &= \rbrac{\th_\l - \f{C_\l}{C_\g}\ \th_\g} \ldot + \rbrac{\th_t - \f{C_t}{C_\g} \th_\g}\\
        &=: \hthl \ldot + \htht
    }.
    \label{eq:thdot_data_combined}
}
This indicates that $\th = \th(\l, t)$ is a surface parameterized by $\l$ and $t$, equipped with a basis of tangent plane $(\hthl, \htht)$.
\section{Optimally transporting the data distribution}
\label{s:app:ot}

We first give a brief description of the theory of optimal transportation. The optimal transport map between the source task and the target task will be used to define a dynamical process for the task. We only compute the transport for the inputs $x$ between the source and target distributions and use a heuristic to obtain the transport for the labels $y$. This choice is made only to simplify the exposition; it is straightforward to handle the case of transport on the joint distribution $p(x,y)$.

If i.i.d samples from the source task are denoted by $\cbrac{x^s_1, \ldots, x^s_{n_s}}$ and those of the target distribution are $\cbrac{x^t_1, \ldots, x^t_{n_t}}$ the empirical source and target distributions can be written as
\[
    p^s(x) = \f{1}{n_s} \sum_{i=1}^{n_s} \delta_{x-x^s_i}, \trm{and}\ p^t(x) = \f{1}{n_t} \sum_{i=1}^{n_t} \delta_{x-x^t_i}
\]
respectively; here $\delta_{x-x'}$ is a Dirac delta distribution at $x'$. Since the empirical data distribution is a sum of a finite number of Dirac measures, this is a discrete optimal transport problem and easy to solve. We can use the Kantorovich relaxation to denote by $\BB$ the set of probabilistic couplings between the two distributions:
\[
    \BB = \cbrac{\G \in \reals_+^{n_s \times n_t}:\ \G \ind_{n_s} = p, \G^\top \ind_{n_s} = q}
\]
where $\ind_{n}$ is an $n$-dimensional vector of ones. The Kantorovich formulation solves for
\beq{
    \G^* = \argmin_{\G \in \BB}\ \sum_{i=1}^{n_s} \sum_{t=1}^{n_t} \G_{ij}\ \k_{ij}
    \label{eq:ot}
}
where $\k \in \reals_+^{n_s \times n_t}$ is a cost function that models transporting the datum $x_i^s$ to $x_j^t$. This is the metric of the underlying data domain and one may choose any reasonable metric for $\k = \norm{x_i^s - x_j^t}_2^2$. The problem in~\cref{eq:ot} is a convex optimization problem and can be solved easily; in practice we use the Sinkhorn's algorithm~\citep{cuturi2013sinkhorn} which adds an entropic regularizer $-h(\G) = \sum_{ij} \G_{ij} \log \G_{ij}$ to the objective in~\cref{eq:ot}.

\subsection{Changing the data distribution}
\label{ss:adapting_data_distribution}

Given the optimal probabilistic coupling $\G^*$ between the source and the target data distributions, we can interpolate between them at any $t \in [0,1]$ by following the geodesics of the Wasserstein metric
\[
    p(x,t) = \argmin_p\ (1-t) \ww^2(p^s, p) + t \ww^2(p, p^t).
\]
For discrete optimal transport problems, as shown in~\citet{villani2008optimal}, the interpolated distribution $p_t$ for the metric $\k_{ij} = \norm{x_i^2 - x_j^t}^2_2$ is given by
\beq{
    p(x,t) = \sum_{i=1}^{n_s} \sum_{j=1}^{n_t}\ \G^*_{ij}\ \delta_{x-(1-t) x_i^s - t x_j^t}.
    \label{eq:interp_p_ot}
}
Observe that the interpolated data distribution equals the source and target distribution at $t=0$ and $t=1$ respectively and it consists of linear interpolations of the data in between.

\begin{remark}[Interpolating the labels]
\label{rem:interp_labels}
The interpolation in~\cref{eq:interp_p_ot} gives the marginal on the input space interpolated between the source and target tasks. To evaluate the functionals in~\cref{s:dynamical_processes_equilibrium_surface} for the classification setting, we would also like to interpolate the labels. We do so by setting the true label of the interpolated datum $x = (1-t) x_i^s + t x_j^t$ to be linear interpolation between the source label and the target label.
\[
    y(x, t) = (1-t) \delta_{y-y_{x^s_i}} + t \delta_{y-y_{x^t_j}}
\]
for all $i, j$. Notice that the interpolated distribution $p(x,t)$ is a sum of Dirac delta distributions weighted by the optimal coupling. We therefore only need to evaluate the labels at all the interpolated data.
\end{remark}

\begin{remark}[Linear interpolation of data]
\label{rem:linear_interpolation_of_data}
Our formulation of optimal transportation leads to a linear interpolation of the data in~\cref{eq:interp_p}. This may not work well for image-based data where the square metric $\k_{ij} = \norm{x_i^s - x-k^t}_2^2$ may not be the appropriate metric. We note that this interpolation of data is an artifact of our choice of $\k_{ij}$, other choices for the metric also fit into the formulation and should be viable alternatives if they result in efficient computation.
\end{remark}

\section{Transfer learning between two subsets of CIFAR-10}
\label{s:app:transfer_cifar_heuristic}

The iso-classification process is a quasi-static process, i.e., the model parameters $\th$ are lie on the equilibrium surface at all times $t \in [0,1]$ during the transfer. Note that both the equilibrium surface and the free-energy $\ff(\l, \g)$ are functions of the data and change with time. Let us write this explicitly as
\[
    F(t) := R(t, \l(t), \g(t)) + \l D(t, \l(t), \g(t)) + \g C_0
\]
where $C_0$ is the classification loss. We prescribed a geodesic transfer above where the Lagrange multipliers $\l, \g$ were adapted simultaneously to confirm to the constraints of the equilibrium surface locally. We can forgot this and instead adapt them using the following heuristic. We let $\ldot = k$ for some constant $k$ and use
\beq{
    \pf[C]{\l} \ldot + \pf[C]{\g} \gdot + \pf[C]{t} = 0,
    \label{eq:iso_c_trans_system}
}
to get the evolution curve of $\g(t)$.

\begin{figure}[!h]
\centering
    \begin{subfigure}[b]{0.35\textwidth}
    \centering
    \includegraphics[width=\textwidth]{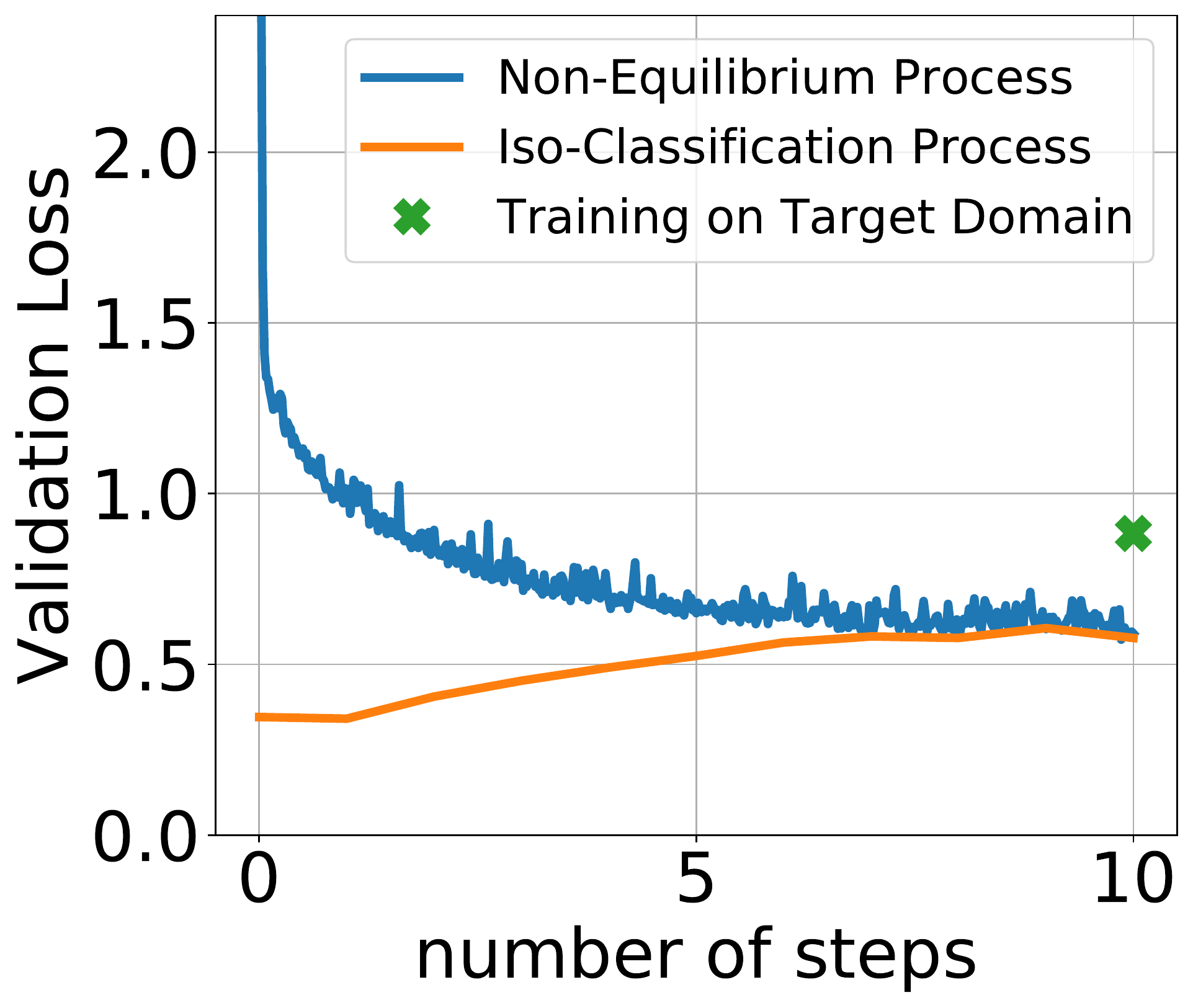}
    \caption{}
    \label{fig:cifar_transfer:c}
    \end{subfigure}
    \hspace{0.25in}
    \begin{subfigure}[b]{0.35\textwidth}
    \centering
    \includegraphics[width=\textwidth]{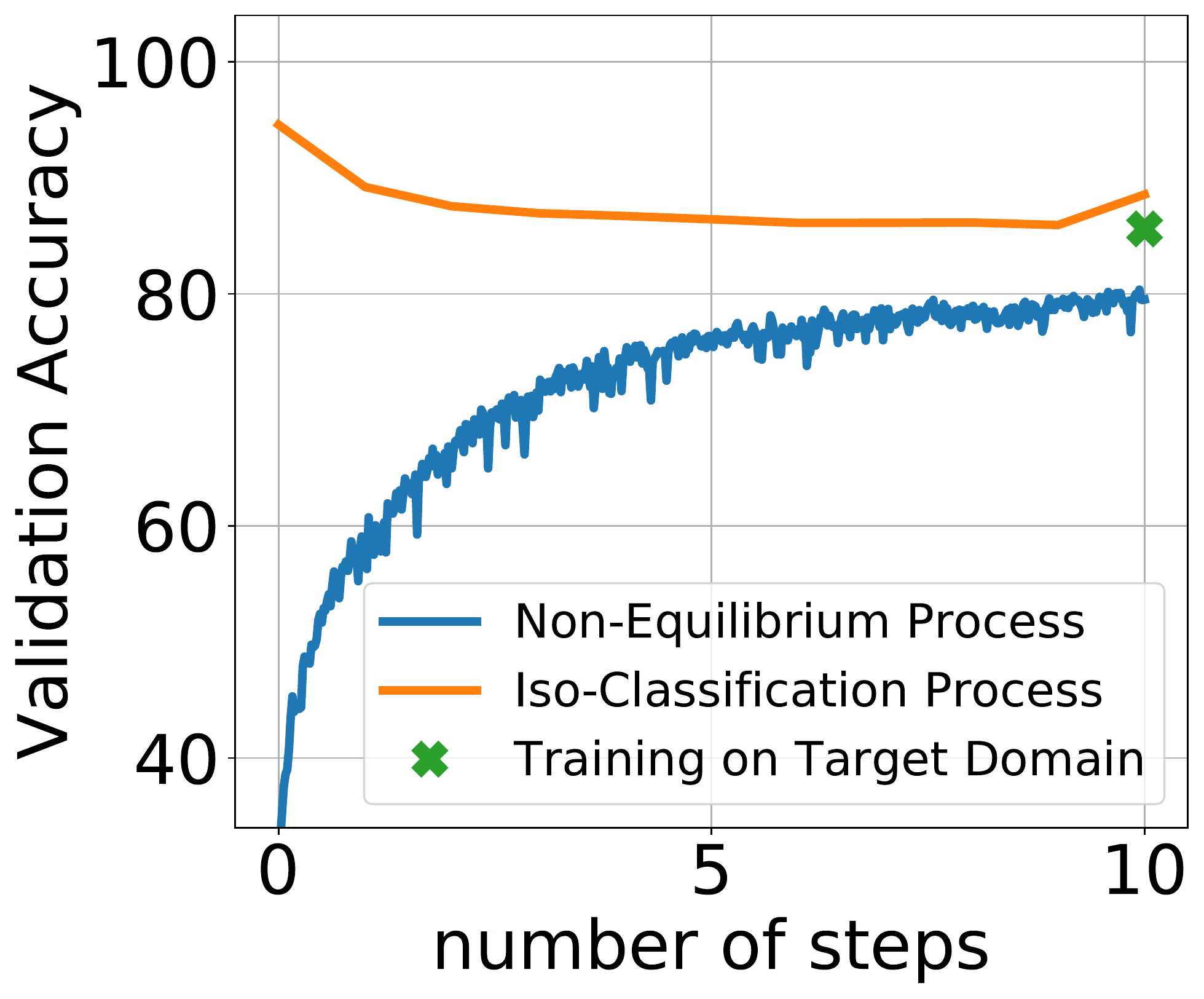}
    \caption{}
    \label{fig:cifar_transfer:acc}
    \end{subfigure}
\vspace*{-0.2in}
\caption{\tbf{Transferring from source dataset of CIFAR-10 vehicles to the target dataset consisting of four animals.} \cref{fig:cifar_transfer:c} shows the variation of validation loss during the transfer. \cref{fig:cifar_transfer:acc} shows the validation accuracy during the transfer. The orange curve corresponds to iso-classification transfer; the blue curve is the result of directly fine-tuning the source model on the target data (note the very low accuracy at the start); the green point is the accuracy of training on the target task from scratch.}
\label{fig:cifar_transfer}
\end{figure}

Here we present experimental results of an iso-classification process for transferring the learnt representation. We pick the source dataset to be all vehicles (airplane, automobile, ship and truck) in CIFAR-10 and the target dataset consists of four animals (bird, cat, deer and dog). We set the output size of classifier to be four. Our goal is to adapt a model trained on the source task to the target task while keeping its classification loss constant. We run the iso-c transfer dynamics~\cref{eq:iso_c_trans_system}
and the results are shown in~\cref{fig:cifar_transfer}.

It is evident that both the classification accuracy and loss are constant throughout the transfer. CIFAR-10 is a more complex dataset as comparing with MNIST and the accuracy gap between iso-classification transfer, fine-tuning from the source and training from scratch is significant. Observe that the classification loss gap between iso-classification transfer and training from scratch on the target is also significant. The benefit of running the iso-classification transfer is that we can be guaranteed about the final accuracy and validation loss of the model.

\subsection{ Details of the experimental setup for CIFAR transferring}

At moment $t$, parameters $\l$, $\g$ determine our objective functions. We compute iso-classification loss transfer process by first setting initial states: $(\l = 4, \g = 100)$. We train on source dataset for 300 epochs with Adam and a learning rate of 1E-3 that drops by a factor of 10 after every 120 epochs to obtain the initial state. We change $\l$, $\g$ with respect to time $t$ and then apply the equilibration learning rate schedule of~\cref{fig:ff_equilibriation} to achieve the transition between equilibrium states. We compute the partial derivatives $\pf[C]{t}$, $\pf[C]{\l}$ and $\pf[C]{\g}$  by using finite difference. At each time $t$, solving~\cref{eq:iso_c_trans_system} with the partial derivatives leads to the solution for $\gdot$, where $ \dot{\l}$ is a constant. In our experiment we set $\dot{\l} = - 1.5$.

\end{appendix}

\end{document}